\documentclass[10pt,twocolumn,letterpaper]{article}

\usepackage[pagenumbers]{cvpr} 

\usepackage{graphicx}
\usepackage{amsmath}
\usepackage{amssymb}
\usepackage{booktabs}
\usepackage{slashbox}
\usepackage[table,xcdraw]{xcolor}
\usepackage{sidecap}
\usepackage{bbm}
\usepackage{caption, booktabs}
\usepackage{caption}
\usepackage{times}
\usepackage{epsfig}
\usepackage{multirow}
\usepackage{lipsum}
\usepackage{enumitem}
\usepackage{subfiles}

\usepackage[pagebackref,breaklinks,colorlinks]{hyperref}

\usepackage[capitalize]{cleveref}
\crefname{section}{Sec.}{Secs.}
\Crefname{section}{Section}{Sections}
\Crefname{table}{Table}{Tables}
\crefname{table}{Tab.}{Tabs.}

\def\cvprPaperID{9177} 
\def\confName{CVPR}
\def\confYear{2023}

\begin{document}

\title{Bridging Precision and Confidence: \\ A Train-Time Loss for Calibrating Object Detection}

\author{Muhammad Akhtar Munir$^{1, 2}$
~~~
Muhammad Haris Khan$^{1}$
~~~
Salman Khan$^{1,3}$
~~~
Fahad Shahbaz Khan$^{1, 4}$
\\
$^{1}$Mohamed bin Zayed University of AI~~~
$^{2}$Information Technology University~~~\\
$^{3}$Australian National University~~~
$^{4}$Linköping University
}
\maketitle

\begin{abstract}
Deep neural networks (DNNs) have enabled astounding progress in several vision-based problems. Despite showing high predictive accuracy, recently, several works have revealed that they tend to provide overconfident predictions and thus are poorly calibrated. The majority of the works addressing the miscalibration of DNNs fall under the scope of classification and consider only in-domain predictions. However, there is little to no progress in studying the calibration of DNN-based object detection models, which are central to many vision-based safety-critical applications. In this paper, inspired by the train-time calibration methods, we propose a novel auxiliary loss formulation that explicitly aims to align the class confidence of bounding boxes with the accurateness of predictions (i.e. precision). Since the original formulation of our loss depends on the counts of true positives and false positives in a minibatch, we develop a differentiable proxy of our loss that can be used during training with other application-specific loss functions. We perform extensive experiments on challenging in-domain and out-domain scenarios with six benchmark datasets including MS-COCO, Cityscapes, Sim10k, and BDD100k. Our results reveal that our train-time loss surpasses strong calibration baselines in reducing calibration error for both in and out-domain scenarios. 
Our source code and pre-trained models are available at
\url{https://github.com/akhtarvision/bpc_calibration}
   
\end{abstract}
\vspace{-0.15cm}

\begin{figure}[t]
    \centering
    \includegraphics[width=0.9\linewidth]{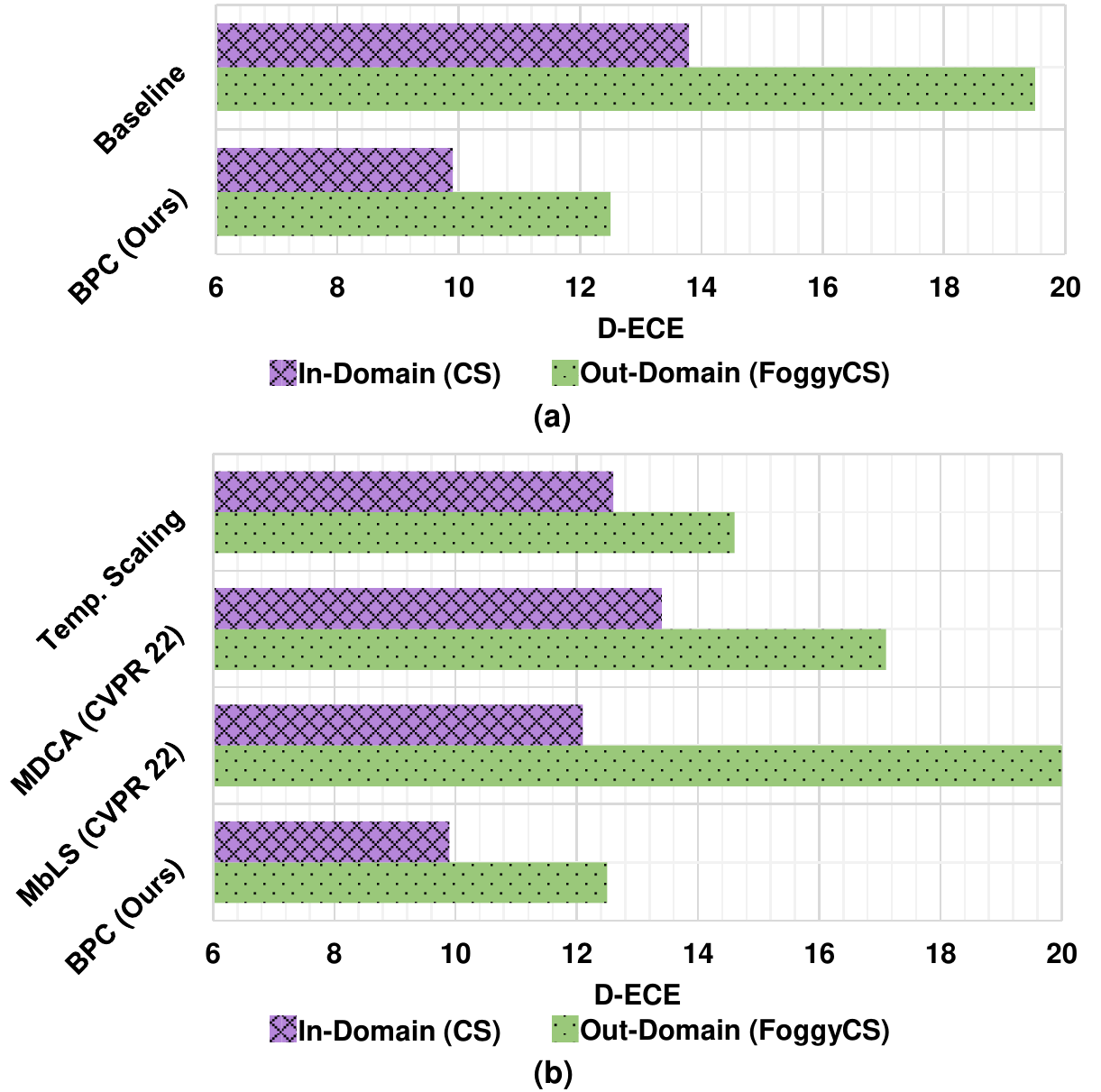}
    \captionsetup{justification=justified}
    \caption{\small Comparison in terms of Detection Expected Calibration Error (D-ECE) on in-domain Cityscapes (CS) and out-domain FoggyCS datasets. (a) Detection model trained with our proposed BPC loss provides the lowest D-ECE (lower is better). 
    (b) Post-hoc and train-time calibration methods for classification: MDCA \cite{hebbalaguppe2022stitch} and MbLS\cite{Liu_2022_CVPR} are sub-optimal for object detection. In contrast, BPC loss better calibrates in-domain and out-domain detections.} 
    \label{fig:teaserCSfoggy}
    \vspace{-0.15cm}
\end{figure}

\section{Introduction}
\label{sec:intro}

Deep neural networks (DNNs) have shown remarkable results in various mainstream computer vision tasks, including image classification \cite{simonyan2014very,he2016deep,dosovitskiy2021an}, object detection \cite{ren2015faster, tian2019fcos, zhu2021deformable}, and semantic segmentation  \cite{chen2018deeplab, Segmenter2021}. 
However, some recent works \cite{guo2017calibration, mukhoti2020calibrating} show that these deep models have the tendency to provide overconfident predictions. This greatly limits the overall trust in their predictions, especially when they are part of the decision-making system in safety-critical applications \cite{grigorescu2020survey,dusenberry2020analyzing, sharma2017crowdsourcing}. For instance, a decision system in an AI-powered healthcare diagnostic application can safely reject predictions with low confidence, however, if it mistakenly skips reviewing an incorrect prediction with high confidence, it can lead to serious consequences.  

An important underlying reason behind the miscalibration of DNNs is training with zero-entropy supervision signal which makes them overconfident, and thus inadvertently miscalibrated. There have been few attempts towards improving the model calibration. A prominent technique is based on a post-processing step that transforms the outputs of a trained model with parameter(s) learned on a held-out validation set \cite{guo2017calibration, islam2021class, Ji2019BinwiseTS, platt1999probabilistic, kull2017beta}. Although simple to implement, these methods are architecture and data-dependent \cite{Liu_2022_CVPR}, and further requires a separate held-out validation set which is not readily available in many real-world applications. An alternative approach is a train-time calibration method which tends to involve all model parameters during training. Existing train-time calibration methods \cite{hebbalaguppe2022stitch, Liu_2022_CVPR, liang2020imporved, krishnan2020improving} propose an auxiliary loss term that can be used in conjunction with an application-specific loss function (e.g., Cross Entropy or Focal loss \cite{mukhoti2020calibrating}). Recently,   \cite{hebbalaguppe2022stitch} propose a differentiable auxiliary loss formulation to calibrate the class confidence of both the predicted label along with non-predicted labels.

\begin{figure*}[t]
    \centering
    \includegraphics[width=1.0\linewidth]{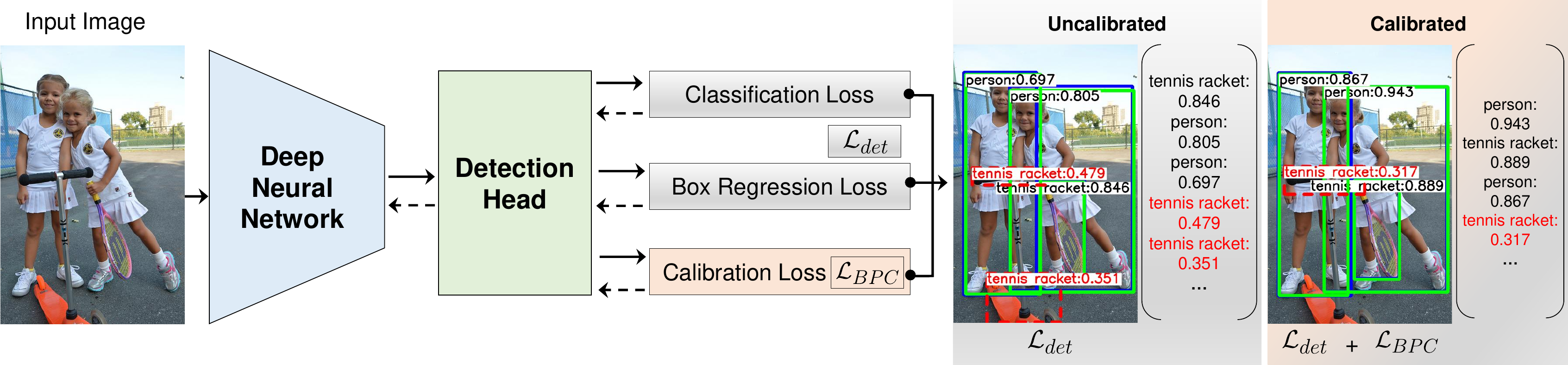}
    \captionsetup{justification=justified}
    \caption{\small Main architecture: Our BPC loss function is integrated with object detection architecture and the detector predicts well-calibrated probabilities for accurate predictions (shown in \textcolor{green}{Green}), while lowering the probabilities of inaccurate prediction (shown in dashed \textcolor{red}{Red}). On the other hand, an uncalibrated model predicts scores, lower for accurate predictions and higher for inaccurate predictions. \textcolor{blue}{Blue} color shows the ground truth boxes present for corresponding detections.
    Best viewed in color.
    }
    \label{fig:arch_diag}
\end{figure*}

Almost all work towards improving model calibration target the task of classification \cite{hebbalaguppe2022stitch, Liu_2022_CVPR, liang2020imporved, guo2017calibration, kumar2018trainable}. However, the calibration of object detection models has not been actively explored. 
Similar to classification models, object detection models also occupy an important position in many safety-critical applications. For instance, they form an integral part of the perception component of self-driving vehicles. Furthermore, the majority of efforts tackling model calibration focus on calibrating in-domain predictions. A deployed deep learning-based model can encounter samples from a distribution that is radically different from the training distribution. Therefore, a real-world model should be well-calibrated for both in-domain and out-domain predictions. So, in essence, well-calibrated object detectors, particularly under distribution shifts, not only contribute to algorithmic advancements but are of great importance to many vision-based safety-critical applications.

In this paper, we study the calibration of object detection models for both in-domain and out-domain predictions.
We observe that the recent state-of-the-art object detectors are rather miscalibrated when compared to their predictive accuracy (Fig.~\ref{fig:teaserCSfoggy}). 
To this end, inspired by the train-time calibration approaches \cite{hebbalaguppe2022stitch, Liu_2022_CVPR, krishnan2020improving}, we propose a novel train-time auxiliary loss formulation (Fig.~\ref{fig:arch_diag}), which explicitly attempts to bridge the model's precision with the predicted class confidence (BPC). It leverages the count of true positives and false positives in a minibatch, which are then employed to construct a penalty for miscalibrated predictions. 
We develop a differentiable proxy to the actual loss formulation that is based on counts. Our loss function is designed to be used with other application-specific loss functions. 
We perform extensive experiments on both in-domain and out-domain scenarios, including the large-scale MS-COCO benchmark. Results reveal that our train-time auxiliary loss is capable of significantly improving the calibration of a state-of-the-art vision-transformer based object detector under both in-domain and out-domain scenarios.

\section{Related Work}
\label{sec:related}

Most of the work for calibrating DNNs can be categorized as: post-hoc and train-time methods. Post-hoc methods require hold-out validation set and involve a few parameters, whereas train-time methods do not require validation data and involve all model parameters. 
We briefly discuss these methods and other works below.


\noindent\textbf{Post-hoc methods:}
A simple and classic approach to improving model calibration is temperature scaling (TS) \cite{guo2017calibration}, which is an extension of Platt scaling \cite{platt1999probabilistic} from binary to multi-class settings. 
TS uses a parameter to modulate the logits of a trained model, whereby this parameter is estimated using hold-out data. This lowers the predicted confidence to achieve calibration. 
A more general form of TS is matrix scaling for the transformation of logits. This  matrix is learned in a similar way using hold-out validation set.
Besides involving limited parameters, the majority of post-hoc methods are limited to calibrating in-domain predictions \cite{ovadia2019can}. Further, these post-hoc calibration methods are prone to performing poorly for dense prediction tasks \cite{hebbalaguppe2022stitch}.
To improve post-hoc calibration under out-domain scenarios, \cite{tomani2021post} transforms the validation set prior to performing the post-hoc approach.
In \cite{ding2021local}, a regression model is used to predict temperature parameter. Post-hoc calibration methods are simple and effective, however, they require hold-out validation data, and are dependent on architecture  \cite{Liu_2022_CVPR}.

\noindent\textbf{Train-time calibration methods:}
Models trained with zero-entropy supervision tend to give over-confident predictions. An example is negative log-likelihood (NLL), which is a widely-used task-specific loss. A model trained with NLL provides predictions that deviate from the accuracy, leaving the model poorly calibrated \cite{guo2017calibration}.
Train-time calibration methods are typically based on auxiliary loss functions, which are used in-tandem with task-specific losses. 
In \cite{liang2020imporved}, an auxiliary loss term DCA is proposed to calibrate the model. It is combined with a task-specific loss to penalize when it reduces but the accuracy remains unchanged.
Likewise, \cite{kumar2018trainable} proposed an auxiliary loss function that is based on a reproducing kernel in a Hilbert space \cite{gretton2013introduction}. \cite{krishnan2020improving} calibrated uncertainty based on the relationship between accuracy and uncertainty. 
Recently in \cite{hebbalaguppe2022stitch}, proposed a loss known as the multi-class difference of confidence and accuracy which aims to calibrate the predicted confidence of all classes.
Building on the label smoothing (LS) work \cite{szegedy2016rethinking}, \cite{Liu_2022_CVPR}, introduced a margin constraint logit distances to achieve implicit model calibration.

\noindent\textbf{Other methods:}
Model calibration with OOD detection in \cite{hein2019relu} suggested that the ReLU activation function causes the model to provide overconfident predictions for input samples that lie away from the training samples. To circumvent this, a model is forced to output low scores for samples distant from training data by leveraging data augmentation using adversarial training.
In \cite{karimi2022improving}, OOD inputs are detected with spectral analysis over early layers in convolutional neural networks (CNNs), thereby achieving model calibration.

All post-hoc and train-time losses target the calibration of classification models, and there is almost no attention given to the calibration of object detection models.
To this end, we explore the space of calibrating modern DNN-based object detectors. We propose a new train-time calibration method based on a new auxiliary loss function (BPC). It is differentiable, operates over mini-batches, and effectively calibrates modern object detectors for in-domain and out-domain detections.


\section{Method}
\label{sec:method}

\subsection{What is calibration?} 
A model is well-calibrated when the predicted confidence is aligned with the likelihood of the sample being correct. 
For example, a prediction of a calibrated model with confidence $s$ aligns with the occurrence of a sample with the same $s$.
A model is overconfident when it satisfies the condition of correctness with $<s\%$, and underconfident when $>s\%$. 
Many recent works addressing model calibration target the task of classification. In the following, we briefly define calibration for classification and object detection.

\noindent\textbf{Classification:}
Given a dataset $\mathcal{D}$ defined with the joint distribution $\mathcal{D}(\mathcal{X},\mathcal{Y})$ such that $N$ number of images belonging to $C$ ground truth classes are available. Let $\mathcal{D} = \{(\mathbf{x}_{n},y_{n})_{n=1}^{N}\}$, where $\mathbf{x}_{n} \in \mathbb{R}^{H \times W \times d}$ (an input image with height $H$, width $W$, and number of channels $d$). For each image, we have a corresponding ground truth class label $y_{n} \in \mathcal{Y} = \{1,2,...,C\}$. Let $\mathcal{G}_{cls}$ be a classification model that predicts a label $\bar{y}$ with confidence score $\bar{s}$. Following \cite{guo2017calibration}, we define a perfect classification calibration as:
$\mathbb{P}(\bar{y}= y|\bar{s}=s) = s, ~\text{s.t., } s \in [0,1]$.
According to this expression, accuracy and confidence must align for all confidence levels.

\noindent\textbf{Object Detection:}
For object detection, the localization of an object is an integral component along with the class label.
Therefore, bounding box ($\mathbf{b}$) annotations are also available with corresponding class labels.
Specifically, for each object, let $\mathbf{b} \in \mathbb{R}^4$ and $y$ correspond to its class label. 
Given the object detector $\mathcal{G}_{det}$, model predicts a bounding box $\mathbf{\bar{b}}$ and label $\bar{y}$ with confidence score $\bar{s}$. Following \cite{kuppers2020multivariate}, we can define the perfect calibration in object detection as:
$\mathbb{P}(K = 1|\bar{s}=s) = s, ~ \forall s \in [0,1]$ \footnote{Note that, this definition of calibration for object detectors is extendable to include box properties.}.
Where $K=1$ denotes an accurate detection in which both the class prediction matches with the ground truth class and the IoU (between the predicted and the ground truth box) is greater than a certain threshold i.e. $\mathbbm{1}[IoU(\mathbf{\bar{b}}, \mathbf{b}) \geq \rho]\mathbbm{1}[\bar{y}= y]$.

\subsection{Measuring Calibration}
\noindent\textbf{Classification:} Expected calibration error (ECE) is a widely used metric to quantify the miscalibration of a classification model. It measures the expected deviation of accuracy from the confidence for all confidence levels \cite{guo2017calibration}.

\begin{equation}
    \mathbb{E}_{\bar{s}}\Big[|\mathbb{P}(\bar{y}= y|\bar{s}=s) - s|\Big]
\end{equation}

As the confidence score is a continuous random variable, the confidence levels are divided into $L$ equally-spaced bins. The approximation of ECE is computed as:
\begin{equation}
    \mathbf{ECE} = \sum_{l=1}^{L} \frac{|B(l)|}{|\mathcal{D}|}\left|\mathrm{acc}(l) - \mathrm{conf}(l)\right|
    \label{Eq:ECE_formula}
\end{equation}
where $|\mathcal{D}|$ is the total number of samples, $B(l)$ is the set of samples in the $l^{th}$ bin. Further, $\mathrm{acc}(l)$ and $\mathrm{conf}(l)$ denote the average accuracy and average confidence over samples in the $l^{th}$ bin, respectively.

\noindent\textbf{Object detection:}
Similar to ECE for classification, we can define the detection expected calibration error (D-ECE) as the expected deviation of precision from the confidence for all confidence levels \cite{kuppers2020multivariate}:

\begin{equation}
    \mathbb{E}_{\bar{s}}\Big[|\mathbb{P}(K=1|\bar{s}=s) - s|\Big]
\end{equation}

As confidence is a continuous variable, similar to eq.(\ref{Eq:ECE_formula}), we can approximate D-ECE:

\begin{equation}
    \mathbf{D\text{-}ECE} = \sum_{l=1}^{L} \frac{|B(l)|}{|\mathcal{D}|}\left|\mathrm{prec}(l) - \mathrm{conf}(l)\right|
    \label{Eq:D-ECE_formula}
\end{equation}
where $\mathrm{prec}(l)$ denotes the precision in $l^{th}$ bin. Different from Eq.~(\ref{Eq:ECE_formula}), here, $B(l)$ is the set of object instances in $l^{th}$ bin and $|\mathcal{D}|$ is the total number of object instances.

\begin{figure}[t]
    \centering
    \includegraphics[width=0.9\linewidth]{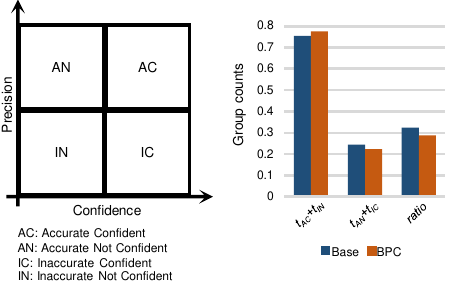}
    \captionsetup{justification=justified}
    \caption{\small Left: We divide the confidence and precision space into four groups for categorizing accurate and inaccurate predictions over the minibatch according to their predicting confidence.
    Right: Compared to baseline, our BPC loss function has increased the joint count of accurate/confident and inaccurate/not-confident detections, while at the same time, it has decreased the joint count of accurate/not-confident and inaccurate/confident detections. Further, when compared to the baseline, the ratio between the latter and the former (defined in Eq. (\ref{eq:mainloss})) is smaller.} 
    \label{fig:group_diag}
\end{figure}

\subsection{BPC: Train-time Calibration Loss for Detection}
\label{subsection:BPC Calibration Loss}

\noindent\textbf{Motivation:}
DNNs-based object detectors are trained with the objective to predict with high confidence, leaving them miscalibrated for both in-domain and out-of-domain detections.
The rationale behind this behavior is the lack of direct supervision for the model to promote higher confidence for accurate predictions and lower confidence for inaccurate predictions.
Motivated by this observation, we leverage the statistics associated with high-scoring and low-scoring box predictions to calibrate the detection model. 
We utilize the true positives and false positives to span the precision and confidence space in order to maximize the probability scores for accurate predictions and minimize the same for inaccurate predictions.
Specifically, we discretize the confidence and precision space into four partitions for categorizing the accurate and inaccurate detections (Fig. \ref{fig:group_diag}).
This lets us develop a simple auxiliary objective, which attempts to distribute the detections according to their respective confidences.

\noindent\textbf{Formulation:}
We propose a train-time method for calibrating object detectors, at the core of which is a simple auxiliary loss function. It is differentiable, operates on minibatches, and is formulated to be used with other task-specific detection losses.  

Inspired by the train-time calibration loss for classification \cite{krishnan2020improving}, we formulate a loss function specific to object detection.
We divide the confidence and precision space into four partitions and categorize the true positive (TP) and false positive (FP) detections over a minibatch. The four partitions for TP and FP are: \textbf{(1)} accurate and confident (AC) \textbf{(2)} accurate and not confident (AN) \textbf{(3)} inaccurate and confident (IC) and \textbf{(4)} inaccurate and not confident (IN). Let $\text{t}_{AC}$, $\text{t}_{AN}$, $\text{t}_{IC}$ and $\text{t}_{IN}$ represent the number of detections in AC, AN, IC, and IN, respectively. In principle, we need accurate detections to be more confident and inaccurate ones to be less confident, so we define the following objective that should be maximized:

\begin{align}
    \text{PC} = \frac{\text{t}_{AC}+\text{t}_{IN}}{\text{t}_{AC}+\text{t}_{IN}+\text{t}_{AN}+\text{t}_{IC}}
    \label{eq:nondift}
\end{align}

In object detection, the obtained predictions are either accurate or inaccurate. Given the predicted class label, bounding boxes, $\mathbbm{1}$ as an indicator function, and $th$ is the threshold on score, we define the following:

\begin{align}
    \text{t}_{AC} = \sum_i \mathbbm{1}[IoU(\mathbf{\bar{b}}_i, \mathbf{b}_i) \geq \rho]\mathbbm{1}[\bar{y}_i= y_i] ~ \mid ~ \bar{s}_i \ge th
    \label{eq:ACeq}
\end{align}
\begin{align}
    \text{t}_{AN} = \sum_i \mathbbm{1}[IoU(\mathbf{\bar{b}}_i, \mathbf{b}_i) \geq \rho]\mathbbm{1}[\bar{y}_i= y_i] ~ \mid ~ \bar{s}_i < th
    \label{eq:ANeq}
\end{align}

\noindent\text{$\text{t}_{IC}$ \& $\text{t}_{IN}$: } The remaining detections after populating $\text{t}_{AC}$ and $\text{t}_{AN}$ are false positives (inaccurate). Similar to Eq. (\ref{eq:ACeq}) and Eq. (\ref{eq:ANeq}), we categorize them based on their confidence scores.

In our loss formulation, we consider precision since it includes true positives and false positives, for which we have confidence scores. Whereas false negatives cannot be considered as they do not have confidence scores because of no detections.
Since Eq. (\ref{eq:nondift}) is not differentiable owing to the indicator functions for $\text{t}_{AC}$, $\text{t}_{AN}$, $\text{t}_{IC}$ and $\text{t}_{IN}$, we formulate its differentiable version to approximate these quantities. Let $t_{AC}$, $t_{AN}$, $t_{IC}$ and $t_{IN}$ be the approximations to $\text{t}_{AC}$, $\text{t}_{AN}$, $\text{t}_{IC}$ and $\text{t}_{IN}$, respectively. We express the differentiable formulation as:

\begin{align}
    \hspace{-0.7cm}t_{AC} &= \sum_{i \in \Big(\substack{K_i=1 ~\& \\ \bar{s}_i \ge th}\Big)} \bar{s}_i \odot \tanh (\bar{s}_i) \notag 
\end{align}
\begin{align}
    t_{AN} &= \sum_{i \in \Big(\substack{K_i=1 ~\& \\ \bar{s}_i<th}\Big)} \bar{s}_i \odot (1-\tanh(\bar{s}_i)) \notag
\end{align}
\begin{align}
    \hspace{-0.7cm}t_{IC} &= \sum_{i \in \Big(\substack{K_i=0 ~\& \\ \bar{s}_i \ge th}\Big)} (1-\bar{s}_i) \odot \tanh(\bar{s}_i) \notag
\end{align}
\begin{align}
    t_{IN} &= \sum_{i \in \Big(\substack{K_i=0 ~\& \\ \bar{s}_i<th}\Big)} (1-\bar{s}_i) \odot (1-\tanh(\bar{s}_i)) \notag
\end{align}

This is based on the rationale that when a detection is accurate, the confidence score satisfies to $\bar{s}$ $\rightarrow$ 1, and otherwise $\bar{s}$ $\rightarrow$ 0.
Where $\tanh$ denotes the hyperbolic tangent function that modulates the penalization to the confidence score.  
The  $\tanh$ function tapers off the confidence values in cases where the prediction is accurate so emphasize less on easy cases (well-calibrated) and focus more on the hard cases (not well-calibrated).
Now we define the differentiable surrogate approximation to our auxiliary loss function:
\begin{align}
    \mathcal{L}_{BPC} = - \log \Bigg(\frac{t_{AC}+t_{IN}}{t_{AC}+t_{IN}+t_{AN}+t_{IC}}\Bigg)
    \label{eq:maxloss}
\end{align}

This above relation Eq.(\ref{eq:maxloss}) can be simplified as to minimize the following:

\begin{align}
    \mathcal{L}_{BPC} = \log \Bigg( 1 + \frac{t_{AN}+t_{IC}}{t_{AC}+t_{IN}}\Bigg)
    \label{eq:mainloss}
\end{align}

We note that, the calibration loss $\mathcal{L}_{BPC}$ is model-agnostic and differentiable. and can be integrated with task specific losses of modern object detection methods. Let $\mathcal{L}_{det}$ be the object detection loss that contains classification (\eg Focal Loss) and localization (\eg Generalized IoU \& L1) losses, we can add our $\mathcal{L}_{BPC}$ to it and obtain the total loss as:

\begin{align}
    \mathcal{L}_{total} = \mathcal{L}_{det} + \mathcal{L}_{BPC}
    \label{eq:totalloss}
\end{align}

Since accurate predictions should have high confidence scores, our train-time loss attempts to align higher accuracy with higher confidence scores and vice versa. 
As shown in Fig.~\ref{fig:group_diag}, compared to baseline, it increases the joint count of accurate/confident and inaccurate/not-confident detections, while at the same time, it decreases the joint count of accurate/not-confident and inaccurate/confident detections. Further, when compared to the baseline, the ratio between the latter and the former (defined in Eq. (\ref{eq:mainloss})) is decreased.

\begin{SCtable*}[][t]
\footnotesize
\centering
\renewcommand{\arraystretch}{1.1}
\begin{tabular}{lcccccc}
\hline
\backslashbox{\textbf{Methods}}{\textbf{Scenarios}}     & \multicolumn{3}{c}{\textbf{In-Domain (COCO)}}                                                 & \multicolumn{3}{c}{\textbf{Out-Domain (CorCOCO)}}                                             \\ \hline\hline
\textbf{}                     & \multicolumn{1}{c}{\textbf{D-ECE $\downarrow$}} & \multicolumn{1}{c}{\textbf{AP box}} & \textbf{mAP@0.5} & \multicolumn{1}{c}{\textbf{D-ECE $\downarrow$}} & \multicolumn{1}{c}{\textbf{AP box}} & \textbf{mAP@0.5} \\ 
\textbf{Baseline \cite{zhu2021deformable}}             & \multicolumn{1}{c}{12.8}           & \multicolumn{1}{c}{44.0}            & 62.9             & \multicolumn{1}{c}{10.8}           & \multicolumn{1}{c}{23.9}            & 35.8             \\ 
\textbf{TS (post-hoc) \cite{guo2017calibration}} & \multicolumn{1}{c}{14.2}           & \multicolumn{1}{c}{44.0}            & 62.9             & \multicolumn{1}{c}{12.3}           & \multicolumn{1}{c}{23.9}            & 35.8             \\ 
\textbf{MDCA   \cite{hebbalaguppe2022stitch}}     & \multicolumn{1}{c}{12.2}           & \multicolumn{1}{c}{44.0}            & 62.9             & \multicolumn{1}{c}{11.1}           & \multicolumn{1}{c}{23.5}            & 35.3             \\ 
\textbf{MbLS \cite{Liu_2022_CVPR}}     & \multicolumn{1}{c}{15.7}           & \multicolumn{1}{c}{44.4}            & 63.4             & \multicolumn{1}{c}{12.4}           & \multicolumn{1}{c}{23.5}            & 35.3             \\ 
\hline
\textbf{\cellcolor[HTML]{D4F7D4}BPC   (Ours)}        & \multicolumn{1}{c}{\cellcolor[HTML]{D4F7D4}10.3}           & \multicolumn{1}{c}{\cellcolor[HTML]{D4F7D4}43.7}            & \cellcolor[HTML]{D4F7D4}62.8             & \multicolumn{1}{c}{\cellcolor[HTML]{D4F7D4}9.4}            & \multicolumn{1}{c}{\cellcolor[HTML]{D4F7D4}23.2}            & \cellcolor[HTML]{D4F7D4}\cellcolor[HTML]{D4F7D4}34.9             \\ \hline
\end{tabular}
\captionsetup{justification=justified}
\captionof{table}{\small Calibration performance on COCO in-domain and out-domain scenarios. Results show that our proposed BPC improves calibration of object detection as compared to baseline, other train time losses and post-hoc methods. AP box and mAP@0.5 are also reported in the table.}
\label{tab:cocoinout}
\end{SCtable*}

\section{Experiments \& Results}
\label{sec:experiment}

\noindent\textbf{Datasets:}
For both in-domain and out-domain scenarios, we perform experiments on various object detection datasets, including large-scale ones. \textbf{MS-COCO} \cite{lin2014microsoft} contains 118K images for training as train2017, 41K as test2017, and 5K images as val2017, that are used for evaluation. It consists of 80 object categories in real world images. 
\textbf{CorCOCO} is a corrupted version of MS-COCO val2017 dataset for evaluations in out-domain scenarios. It incorporates random corruptions out of specified settings in \cite{hendrycks2019benchmarking}, with arbitrary severity levels.
\textbf{Cityscapes} \cite{Cordts2016Cityscapes} is an urban driving scene dataset consisting of 8 categories: \textit{person, rider, car, truck, bus, train, motorbike, and bicycle}.
It contains 2975 training images and 500 validation images used for evaluation. 
\textbf{Foggy Cityscapes} \cite{sakaridis2018semantic} consists of images simulating foggy weather on Cityscapes, and its validation set with severe level of fog is used for evaluation for out-domain scenario.
\textbf{Sim10k} \cite{johnson2017driving} is a dataset of synthetic images containing car category. It contains 10K images from which we split 8K as training set and 1K is used for evaluation.
\textbf{BDD100k} \cite{yu2020bdd100k} consists of 70K training images, 20K test images and 10K validation images. We only consider daylight subset of validation set for the evaluation of out-domain scenario which counts to 5.2K images. This dataset contains class categories similar to Cityscapes. 

\noindent\textbf{Datasets (post-hoc):} We use validation sets based on three in-domain scenarios for temperature scaling as a post-hoc method. We opt Object365 \cite{Objects3652019} validation dataset in case of MS-COCO with similar categories, subset of BDD100k train set for Cityscapes and for Sim10k, its validation split.

\noindent\textbf{Implementation Details:} We use a SoTA detector Deformable-DETR (D-DETR) as a baseline and integrate our loss function with it. D-DETR uses focal loss \cite{lin2017focal} for classification and generalized IOU \& L1 losses \cite{DETR2020} for localization.
Default settings are used and more details can be seen in \cite{zhu2021deformable}.
In addition to the comparison of our proposed train-time loss with post-hoc method \cite{guo2017calibration}, we also compare calibration performance with recent calibration losses, MDCA \cite{hebbalaguppe2022stitch} and MbLS \cite{Liu_2022_CVPR}. D-DETR is trained with respective train-time losses and in-domain datasets.

\noindent\textbf{Evaluation:} For both in-domain and out-domain, we report detection expected calibration error (D-ECE) \cite{kuppers2020multivariate} as object detection calibration measure along with mean average precision of detectors.

\begin{figure}[t]
    \centering
    \includegraphics[width=0.9\linewidth]{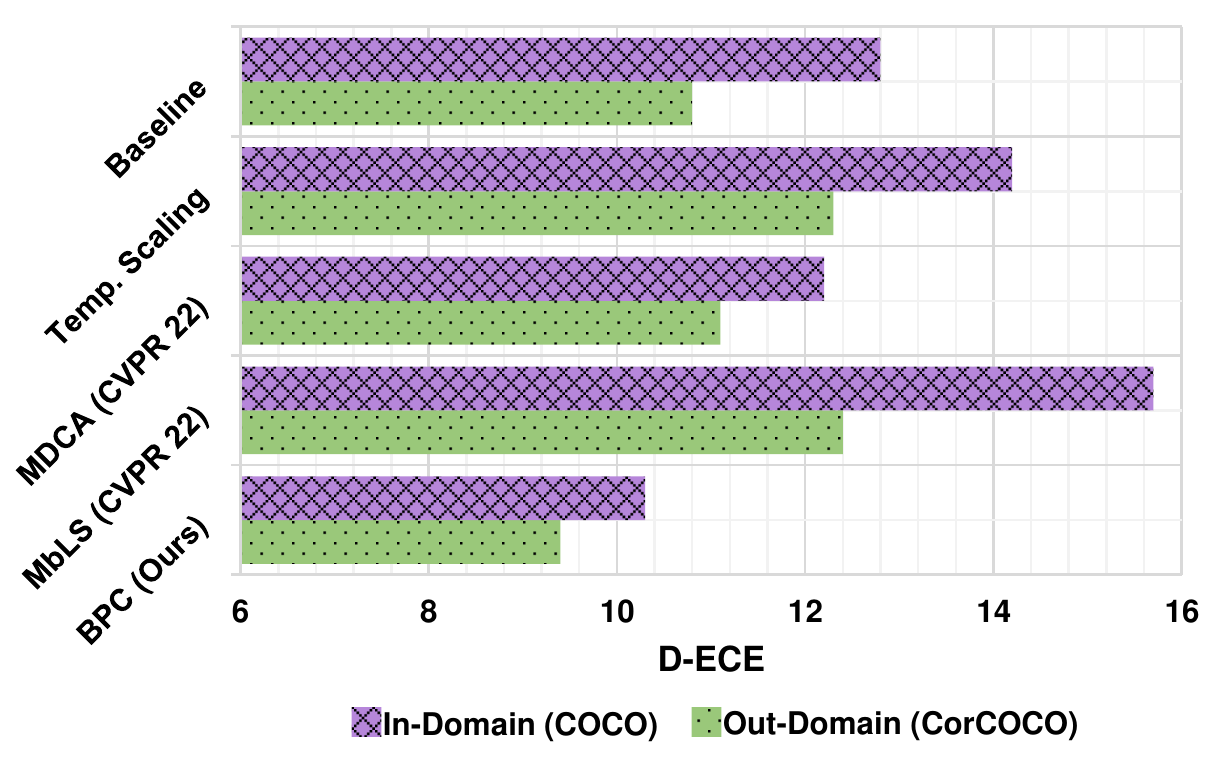}
    \captionsetup{justification=justified}
    \caption{\small Our BPC loss reduces D-ECE for MS-COCO in-domain and Cor-COCO out-domain. We note that the classification calibration losses are sub-optimal for detection calibration and BPC loss improves calibration over post-hoc and train-time losses.}
    \label{fig:COCO_indomain_out}
    \vspace{-0.2cm}
\end{figure}

\begin{table*}[t]
\footnotesize
\centering
\renewcommand{\arraystretch}{1.1}
\begin{tabular}{lccccccccc}
\hline
\backslashbox{\textbf{Methods}}{\textbf{Scenarios}}        & \multicolumn{3}{c}{\textbf{In-Domain (Cityscapes)}}                                                        & \multicolumn{3}{c}{\textbf{Out-Domain (Foggy Cityscapes)}}                                    & \multicolumn{3}{c}{\textbf{Out-Domain (BDD100k)}}                                             \\ 
\hline\hline
\textbf{}                          & \multicolumn{1}{c}{\textbf{D-ECE $\downarrow$}} & \multicolumn{1}{c}{\textbf{AP box}} & \textbf{mAP@0.5} & \multicolumn{1}{c}{\textbf{D-ECE $\downarrow$}} & \multicolumn{1}{c}{\textbf{AP box}} & \textbf{mAP@0.5} & \multicolumn{1}{c}{\textbf{D-ECE $\downarrow$}} & \multicolumn{1}{c}{\textbf{AP box}} & \textbf{mAP@0.5} \\ 
\textbf{Baseline \cite{zhu2021deformable}}                  & \multicolumn{1}{c}{13.8}           & \multicolumn{1}{c}{26.8}            & 49.5             & \multicolumn{1}{c}{19.5}           & \multicolumn{1}{c}{17.3}            & 29.3             & \multicolumn{1}{c}{11.7}           & \multicolumn{1}{c}{10.2}            & 21.9             \\ 
\textbf{TS (post-hoc) \cite{guo2017calibration}} & \multicolumn{1}{c}{12.6}           & \multicolumn{1}{c}{26.8}            & 49.5             & \multicolumn{1}{c}{14.6}           & \multicolumn{1}{c}{17.3}            & 29.3             & \multicolumn{1}{c}{24.5}           & \multicolumn{1}{c}{10.2}            & 21.9             \\ 

\textbf{MDCA \cite{hebbalaguppe2022stitch}} & \multicolumn{1}{c}{13.4}           & \multicolumn{1}{c}{27.5}            & 49.5             & \multicolumn{1}{c}{17.1}           & \multicolumn{1}{c}{17.7}            & 30.3             & \multicolumn{1}{c}{14.2}           & \multicolumn{1}{c}{10.7}            & 22.7             \\ 

\textbf{MbLS \cite{Liu_2022_CVPR}} & \multicolumn{1}{c}{12.1}           & \multicolumn{1}{c}{27.3}            & 49.7             & \multicolumn{1}{c}{20.0}           & \multicolumn{1}{c}{17.1}            & 29.1             & \multicolumn{1}{c}{11.6}           & \multicolumn{1}{c}{10.5}            & 22.7             \\ \hline

\textbf{\cellcolor[HTML]{D4F7D4}BPC (Ours)}                      & \multicolumn{1}{c}{\cellcolor[HTML]{D4F7D4}9.9}            & \multicolumn{1}{c}{\cellcolor[HTML]{D4F7D4}26.8}            & \cellcolor[HTML]{D4F7D4}48.7             & \multicolumn{1}{c}{\cellcolor[HTML]{D4F7D4}12.5}           & \multicolumn{1}{c}{\cellcolor[HTML]{D4F7D4}17.7}            & \cellcolor[HTML]{D4F7D4}30.2             & \multicolumn{1}{c}{\cellcolor[HTML]{D4F7D4}10.6}           & \multicolumn{1}{c}{\cellcolor[HTML]{D4F7D4}11.0}              & \cellcolor[HTML]{D4F7D4}23.6             \\ \hline
\end{tabular}
\captionsetup{justification=justified}
\caption{\small Calibration results with baseline, train-time losses and post-hoc methods are reported. BPC shows improvement in detection calibration for all the scenarios of in-domain (Cityscapes) and out-domain (Foggy Cityscapes \& BDD100k). AP box and mAP@0.5 are also reported for each scenario.}
\label{tab:CSfoggyBDDinout}
\end{table*}

\begin{SCtable*}[][t]
\footnotesize
\centering
\renewcommand{\arraystretch}{1.1}
\begin{tabular}{lcccccc}
\hline
\backslashbox{\textbf{Methods}}{\textbf{Scenarios}} & \multicolumn{3}{c}{\textbf{InDomain (Sim10k)}}                                                        & \multicolumn{3}{c}{\textbf{OutDomain (BDD100k)}}                                                       \\ \hline\hline
\textbf{}                  & \multicolumn{1}{c}{\textbf{D-ECE} $\downarrow$} & \multicolumn{1}{c}{\textbf{AP box}} & \textbf{mAP@0.5} & \multicolumn{1}{c}{\textbf{D-ECE} $\downarrow$} & \multicolumn{1}{c}{\textbf{AP box}} & \textbf{mAP@0.5} \\ 
\textbf{Baseline \cite{zhu2021deformable}}          & \multicolumn{1}{c}{10.3}           & \multicolumn{1}{c}{65.9}            & 90.7             & \multicolumn{1}{c}{7.3}            & \multicolumn{1}{c}{23.5}            & 46.6             \\ 
\textbf{TS (post-hoc) \cite{guo2017calibration}}                & \multicolumn{1}{c}{15.7}           & \multicolumn{1}{c}{65.9}            & 90.7             & \multicolumn{1}{c}{10.5}           & \multicolumn{1}{c}{23.5}            & 46.6             \\ 
\textbf{MDCA \cite{hebbalaguppe2022stitch}}              & \multicolumn{1}{c}{10.0}             & \multicolumn{1}{c}{64.8}            & 90.3             & \multicolumn{1}{c}{8.8}            & \multicolumn{1}{c}{22.7}            & 45.7             \\ 
\textbf{MbLS \cite{Liu_2022_CVPR}}              & \multicolumn{1}{c}{22.5}           & \multicolumn{1}{c}{63.8}            & 90.5             & \multicolumn{1}{c}{16.8}           & \multicolumn{1}{c}{23.4}            & 47.4             \\ \hline
\textbf{\cellcolor[HTML]{D4F7D4}BPC (Ours)}              & \multicolumn{1}{c}{\cellcolor[HTML]{D4F7D4}6.1}            & \multicolumn{1}{c}{\cellcolor[HTML]{D4F7D4}65.4}            & \cellcolor[HTML]{D4F7D4}90.5             & \multicolumn{1}{c}{\cellcolor[HTML]{D4F7D4}6.3}            & \multicolumn{1}{c}{\cellcolor[HTML]{D4F7D4}23.4}            & \cellcolor[HTML]{D4F7D4}45.6             \\ \hline
\end{tabular}
\captionsetup{justification=justified}
\caption{\small Calibration performance with our proposed BPC loss is improved over baseline, train-time losses and post-hoc methods for both in-domain (Sim10k) and out-domain (BDD100k). Car class is considered in this scenario for evaluations. AP box and mAP@0.5 are also reported.}
\label{tab:sim10kbddinout}
\end{SCtable*}

\subsection{Results}
We have performed extensive experiments with post-hoc method and recent train-time losses over various in-domain and out-domain scenarios. For post-hoc method, we need validation set and for temperature scaling (TS) we optimize calibration parameter $T$. 
We compare all of these methods with our proposed loss function, specifically designed for object detectors. Our results show significant improvement over calibration scores (\textbf{lower the better}), while having comparable performance in detection accuracy.

\noindent\textbf{Real and Corrupted domains:}
To see the effectiveness of our loss, we perform experiments with a large-scale benchmark dataset, MS-COCO. We show our results for in-domain and out-domain scenarios, note that a corrupted version of the MS-COCO (CorCOCO) validation set is used for out-domain evaluation. In Tab. \ref{tab:cocoinout}, the post-hoc based TS method fails to improve calibration in both domains, that usually considered to be a good performer in the in-domain. Results show train time losses that are designed for classification-based model calibration, are also not ideal for the calibration of object detectors. We report D-ECE, and our proposed loss function shows improvement in calibration scores for both in-domain ($\rightarrow$COCO, 2.5\%$\downarrow$) and out-domain ($\rightarrow$CorCOCO, 1.4\%$\downarrow$) scenarios from the baseline (Fig. \ref{fig:COCO_indomain_out}).
In comparison with MbLS, our loss improves calibration scores of 5.4\%$\downarrow$ and 3.0\%$\downarrow$ for in-domain and out-domain respectively.

\noindent\textbf{Weather domains:}
We consider weather shift scenario for evaluation in both domains. For in-domain Cityscapes (CS) and out-domain Foggy CS, we see in Tab. \ref{tab:CSfoggyBDDinout} 
our loss shows improvement over post-hoc, for both in-domain ($\rightarrow$CS, 2.7\%$\downarrow$) and out-domain ($\rightarrow$Foggy CS, 2.1\%$\downarrow$). Also, we show improvement as compared to train-time losses, notably ($\rightarrow$Foggy CS, 7.5\%$\downarrow$) over MbLS.

\begin{figure}[t]
    \centering
    \includegraphics[width=0.9\linewidth]{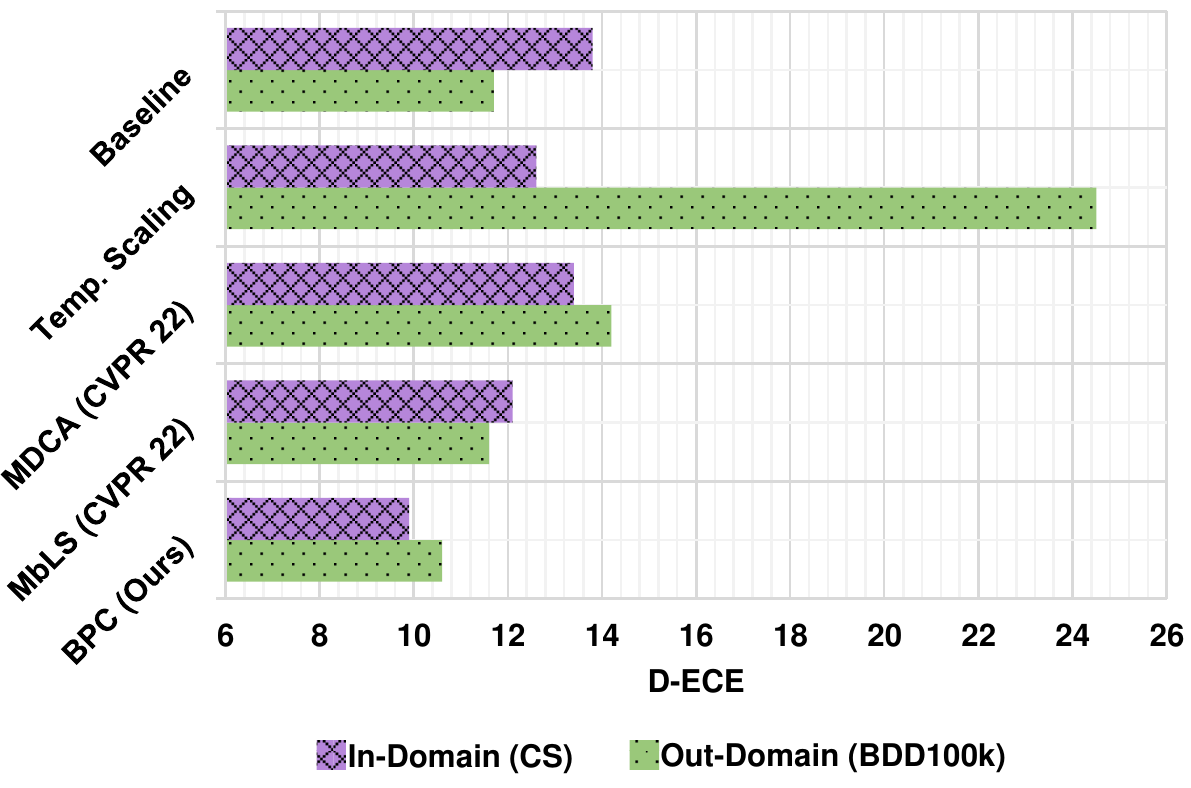}
    \captionsetup{justification=justified}
    \caption{\small After integrating our BPC loss the calibration performance is improved in Cityscapes (CS) and CS to BDD100k.}
    \label{fig:csbdd_idomain_out}
\end{figure}

\noindent\textbf{Scene domains:}
To have CS as in-domain in scene shift, BDD100k is evaluated as an out-domain scenario. Both belong to urban driving scenes but there is a large scene deviation among them. We show results in this scenario in Tab. \ref{tab:CSfoggyBDDinout} and find that TS performs the worst, followed by classification-based train time losses (Fig. \ref{fig:csbdd_idomain_out}). We outperform TS in out-domain ($\rightarrow$BDD100k, 13.9\%$\downarrow$) and the recent MDCA ($\rightarrow$BDD100k, 3.6\%$\downarrow$) approach.

\begin{figure}[t]
    \centering
    \includegraphics[width=0.9\linewidth]{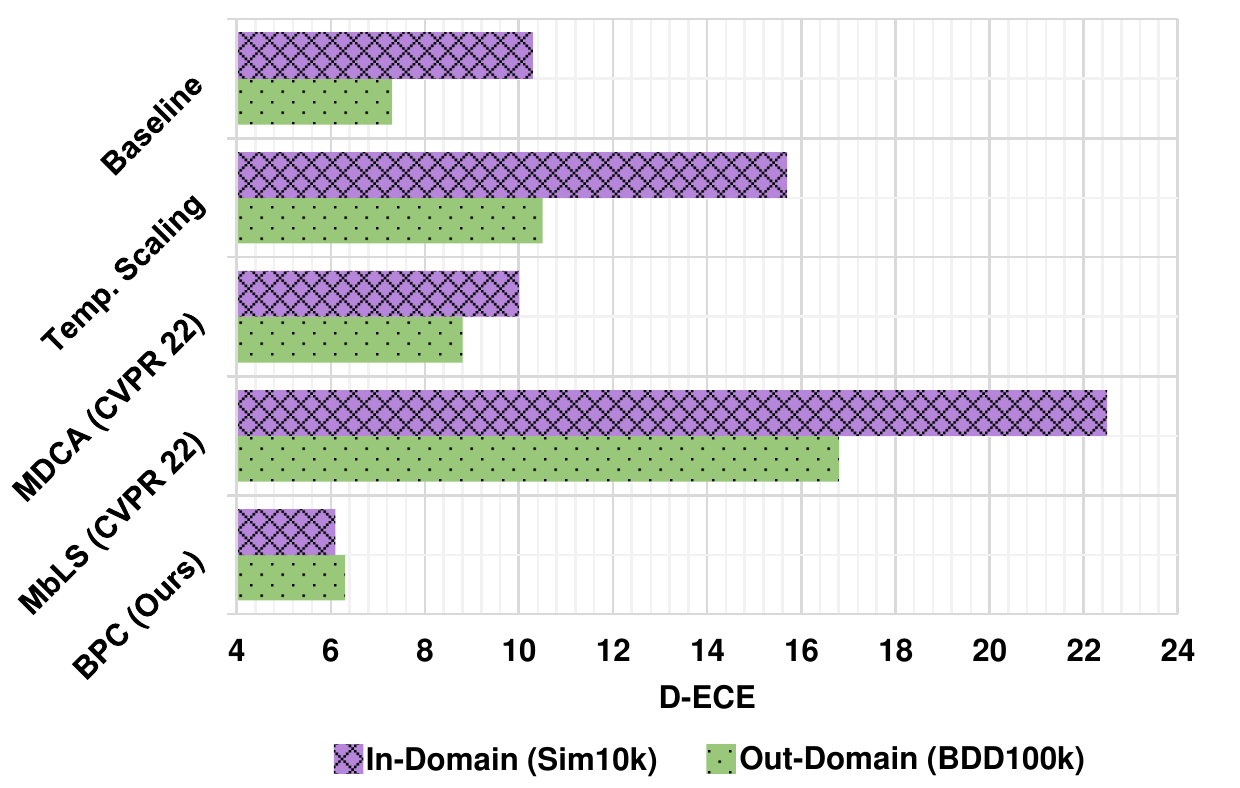}
    \captionsetup{justification=justified}
    \caption{\small Our BPC loss reduces D-ECE in Sim10k and BDD100k as in-domain and out-domain scenarios, respectively.}
    \label{fig:sim10kbdd_idomain_out}
\end{figure}

\noindent\textbf{Synthetic and Real domains:}
Sim10k is a synthetic dataset and considered as in-domain, while BDD100k as a daylight subset is considered as out-domain. 
We extract the car category from the BDD100k evaluation set and report the results. 
Our loss shows improved calibration scores for in-domain ($\rightarrow$Sim10k, 3.9\%$\downarrow$) and out-domain ($\rightarrow$BDD100k, 2.5\%$\downarrow$) scenarios over the MDCA loss (Fig. \ref{fig:sim10kbdd_idomain_out}). 
Also we show calibration improvement of 16.4\%$\downarrow$ and 10.5\%$\downarrow$ over MbLS for in-domain and out-domain respectively (Tab. \ref{tab:sim10kbddinout}).

\begin{figure*}[t]
    \centering
    \includegraphics[width=0.88\linewidth]{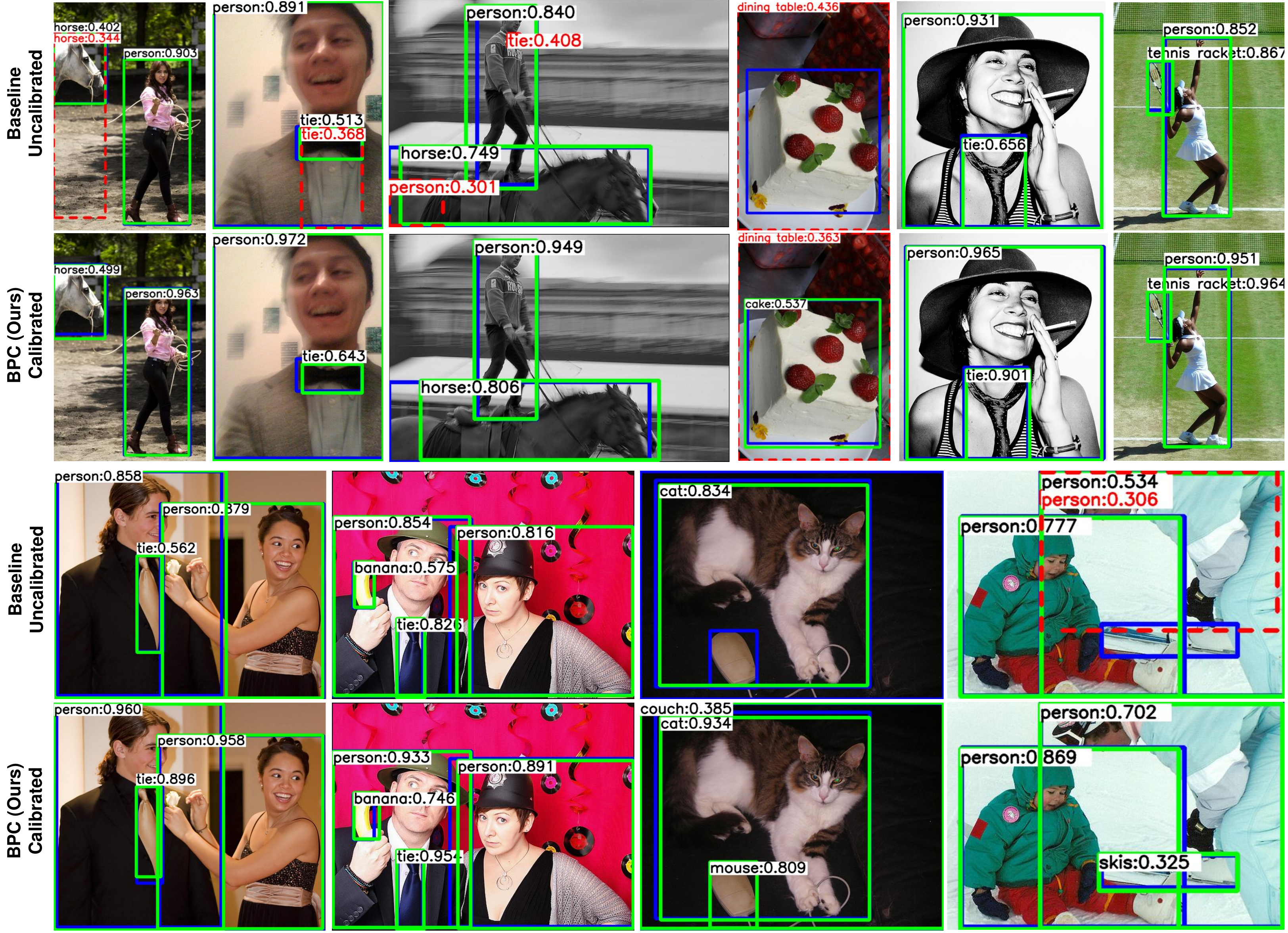}
    \captionsetup{justification=justified}
    \caption{\small Baseline \cite{zhu2021deformable} vs. BPC (Ours): Qualitative results on MS-COCO dataset (In-Domain). Detector trained with our loss forces the accurate predictions to be more confident whereas inaccurate predictions to be less confident. Detection threshold is set to \text{0.3}. \textcolor{green}{Green} boxes are accurate predictions with respective confidence scores. \textcolor{red}{Red} (dashed) boxes are inaccurate predictions with corresponding scores. 
    \textcolor{blue}{Blue} shows the ground truth boxes present for corresponding detections.}
    \label{fig:qual_coco}
\end{figure*}

\noindent\textbf{Qualitative Figures:}
We show qualitative detection results 
in Fig. \ref{fig:qual_coco}. Detector trained with our loss forces the accurate predictions to be more confident whereas inaccurate predictions to be less confident.

\begin{figure*}[t]
    \centering
    \includegraphics[width=1.0\linewidth]{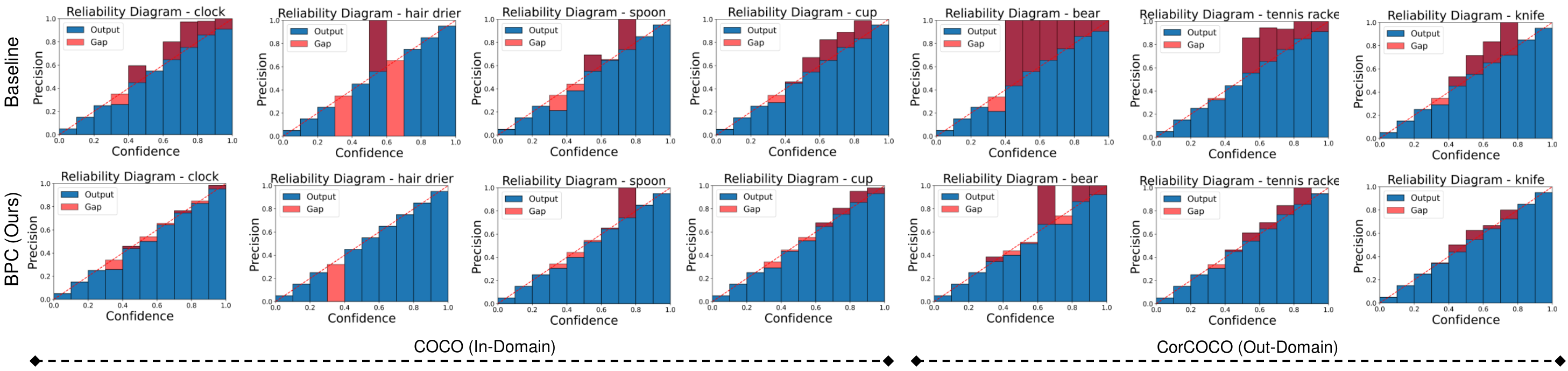}
    \captionsetup{justification=justified}
    \caption{\small Reliability Diagrams: MS-COCO (In-Domain) and MS-COCO (Out-Domain). Top: Baseline \cite{zhu2021deformable} trained as D-DETR and Bottom: D-DETR trained with our proposed BPC loss.}
    \label{fig:RD_qual_coco}
    \vspace{-0.0cm}
\end{figure*}

\noindent\textbf{Reliability Diagrams:}
We show reliability diagrams 
to see the behaviour of calibration in Fig. \ref{fig:RD_qual_coco}. A perfect calibration is achieved if the confidence is exactly same as precision.

\begin{table}[t]
\footnotesize
\centering
\renewcommand{\arraystretch}{1.1}
\begin{tabular}{lccc}
\hline

\textbf{Method}       & \multicolumn{3}{c}{\textbf{In-Domain}}                                                        \\ \hline\hline
                      & \multicolumn{1}{c}{\textbf{D-ECE} $\downarrow$} & \multicolumn{1}{c}{\textbf{AP box}} & \textbf{mAP@0.5} \\ 
\textbf{BPC (th=0.4)} & \multicolumn{1}{c}{9.7}            & \multicolumn{1}{c}{50.2}            & 80.5             \\ 
\textbf{BPC (th=0.5)} & \multicolumn{1}{c}{9.1}            & \multicolumn{1}{c}{50.1}            & 80.2             \\ 
\textbf{BPC (th=0.6)} & \multicolumn{1}{c}{11.2}           & \multicolumn{1}{c}{50.9}            & 81.4             \\ \hline
\end{tabular}
\captionsetup{justification=justified}
\caption{\small Impact of probability thresholds on BPC loss. We perform experiments using train and test subsets of Sim10k train set for ablation study.}
\label{tab:sim10kablprob}
\end{table}

\begin{table}[t]
\footnotesize
\centering
\renewcommand{\arraystretch}{1.1}
\begin{tabular}{lccc}
\hline
\textbf{Method}       & \multicolumn{3}{c}{\textbf{In-Domain}}                                                        \\ \hline\hline
                      & \multicolumn{1}{c}{\textbf{D-ECE} $\downarrow$} & \multicolumn{1}{c}{\textbf{AP box}} & \textbf{mAP@0.5} \\ 
                      \textbf{BPC (BS=1)} & \multicolumn{1}{c}{10.5}            & \multicolumn{1}{c}{50.3}            & 79.6             \\ 
\textbf{BPC (BS=2)} & \multicolumn{1}{c}{9.1}            & \multicolumn{1}{c}{50.1}            & 80.2             \\ 
\textbf{BPC (BS=3)} & \multicolumn{1}{c}{8.9}            & \multicolumn{1}{c}{48.6}            & 78.7             \\ 
\textbf{BPC (BS=4)} & \multicolumn{1}{c}{10.2}           & \multicolumn{1}{c}{47.3}            & 78.3            \\ \hline
\end{tabular}
\captionsetup{justification=justified}
\caption{\small Impact of batch sizes on BPC loss. We observe little degradation in detection accuracy by varying batch size (BS) and observe calibration performance is not much sensitive.}
\label{tab:sim10kablbatch}
\end{table}

\begin{table}[b]
\footnotesize
\centering
\renewcommand{\arraystretch}{1.1}
\begin{tabular}{lccc}
\hline
\textbf{Method}       & \multicolumn{3}{c}{\textbf{In-Domain}}                                                        \\ \hline\hline
                      & \multicolumn{1}{c}{\textbf{D-ECE} $\downarrow$} & \multicolumn{1}{c}{\textbf{AP box}} & \textbf{mAP@0.5} \\ 
\textbf{BPC (seed=30)} & \multicolumn{1}{c}{9.0}            & \multicolumn{1}{c}{49.2}            & 79.7             \\ 
\textbf{BPC (seed=42)} & \multicolumn{1}{c}{9.1}            & \multicolumn{1}{c}{50.1}            & 80.2             \\ 
\textbf{BPC (seed=60)} & \multicolumn{1}{c}{8.6}           & \multicolumn{1}{c}{51.0}            & 80.5            \\ \hline
\end{tabular}
\captionsetup{justification=justified}
\caption{\small Impact of different seeds on BPC loss. We observe changing seeds for initialization has little effect on calibration performance.}
\label{tab:sim10kablseed}
\end{table}

\subsection{Ablation \& Analysis}
We perform ablation studies on score threshold, batch sizes and random initialization. For this purpose, we select the subsets of Sim10k training set as train and validation to empirically find score threshold hyper-parameter. With similar data splits, we show impact of batch sizes and random weight initialization on our loss.

\noindent\textbf{Score Threshold:}
We study the impact of score threshold that is used for penalizing the probabilities of instances present in the batch. Varying the score threshold shows some degradation in detection performance for in-domain but calibration still stands out the best and we find our approach is not much sensitive to it. We empirically find in Tab. \ref{tab:sim10kablprob} that $th=0.5$ improves calibration.

\noindent\textbf{Batch Size:}
We observe in Tab. \ref{tab:sim10kablbatch} the impact of batch sizes on our proposed loss function. We see that increasing batch size has little effect on the detection accuracy and calibration performance is not sensitive for given scenario. To get the best for both metrics and without sacrificing the drop in detection performance, we opt for batch size 2 for all experiments.

\noindent\textbf{Random Weight Initialization:}
Impact of different seeds with calibration loss is studied by setting different initialization points  for experiments. This shows that calibration is not much influenced by random initialization 
(Tab. \ref{tab:sim10kablseed}). We set seed 42 as default in our experiments.

\section{Conclusion}
\label{sec:conclusion}
In this paper, we presented a new train-time calibration method for object detection which is based on an auxiliary loss function (BPC).
It utilizes true positive and false positive statistics to maximize the confidence scores for accurate predictions and minimize the scores for inaccurate predictions.
We perform extensive experiments on several in-domain and out-domain scenarios, including large-scale detection dataset, to show the effectiveness of our loss function for calibrating object detectors.
Results show that our method outperforms several train-time calibration methods in terms of improving the calibration of both in-domain and out-domain predictions while also preserving the detection accuracy.  

{\small
\bibliographystyle{ieee_fullname}
\bibliography{egbib}

\begin{thebibliography}{10}\itemsep=-1pt

\bibitem{DETR2020}
Nicolas Carion, Francisco Massa, Gabriel Synnaeve, Nicolas Usunier, Alexander
  Kirillov, and Sergey Zagoruyko.
\newblock End-to-end object detection with transformers.
\newblock In Andrea Vedaldi, Horst Bischof, Thomas Brox, and Jan-Michael Frahm,
  editors, {\em Computer Vision -- ECCV}, 2020.

\bibitem{chen2018deeplab}
Liang-Chieh Chen, George Papandreou, Iasonas Kokkinos, Kevin Murphy, and Alan~L
  Yuille.
\newblock Deeplab: Semantic image segmentation with deep convolutional nets,
  atrous convolution, and fully connected crfs.
\newblock {\em IEEE transactions on pattern analysis and machine intelligence},
  40(4):834--848, 2018.

\bibitem{Cordts2016Cityscapes}
Marius Cordts, Mohamed Omran, Sebastian Ramos, Timo Rehfeld, Markus Enzweiler,
  Rodrigo Benenson, Uwe Franke, Stefan Roth, and Bernt Schiele.
\newblock The cityscapes dataset for semantic urban scene understanding.
\newblock In {\em Proc. of the IEEE Conference on Computer Vision and Pattern
  Recognition (CVPR)}, 2016.

\bibitem{ding2021local}
Zhipeng Ding, Xu Han, Peirong Liu, and Marc Niethammer.
\newblock Local temperature scaling for probability calibration.
\newblock In {\em Proceedings of the IEEE/CVF International Conference on
  Computer Vision}, pages 6889--6899, 2021.

\bibitem{dosovitskiy2021an}
Alexey Dosovitskiy, Lucas Beyer, Alexander Kolesnikov, Dirk Weissenborn,
  Xiaohua Zhai, Thomas Unterthiner, Mostafa Dehghani, Matthias Minderer, Georg
  Heigold, Sylvain Gelly, Jakob Uszkoreit, and Neil Houlsby.
\newblock An image is worth 16x16 words: Transformers for image recognition at
  scale.
\newblock In {\em International Conference on Learning Representations}, 2021.

\bibitem{dusenberry2020analyzing}
Michael~W Dusenberry, Dustin Tran, Edward Choi, Jonas Kemp, Jeremy Nixon,
  Ghassen Jerfel, Katherine Heller, and Andrew~M Dai.
\newblock Analyzing the role of model uncertainty for electronic health
  records.
\newblock In {\em Proceedings of the ACM Conference on Health, Inference, and
  Learning}, pages 204--213, 2020.

\bibitem{gretton2013introduction}
Arthur Gretton.
\newblock Introduction to rkhs, and some simple kernel algorithms.
\newblock {\em Adv. Top. Mach. Learn. Lecture Conducted from University College
  London}, 16:5--3, 2013.

\bibitem{grigorescu2020survey}
Sorin Grigorescu, Bogdan Trasnea, Tiberiu Cocias, and Gigel Macesanu.
\newblock A survey of deep learning techniques for autonomous driving.
\newblock {\em Journal of Field Robotics}, 37(3):362--386, 2020.

\bibitem{guo2017calibration}
Chuan Guo, Geoff Pleiss, Yu Sun, and Kilian~Q Weinberger.
\newblock On calibration of modern neural networks.
\newblock In {\em International Conference on Machine Learning}, pages
  1321--1330. PMLR, 2017.

\bibitem{he2016deep}
Kaiming He, Xiangyu Zhang, Shaoqing Ren, and Jian Sun.
\newblock Deep residual learning for image recognition.
\newblock In {\em Proceedings of the IEEE conference on computer vision and
  pattern recognition}, pages 770--778, 2016.

\bibitem{hebbalaguppe2022stitch}
Ramya Hebbalaguppe, Jatin Prakash, Neelabh Madan, and Chetan Arora.
\newblock A stitch in time saves nine: A train-time regularizing loss for
  improved neural network calibration.
\newblock In {\em Proceedings of the IEEE/CVF Conference on Computer Vision and
  Pattern Recognition (CVPR)}, pages 16081--16090, June 2022.

\bibitem{hein2019relu}
Matthias Hein, Maksym Andriushchenko, and Julian Bitterwolf.
\newblock Why relu networks yield high-confidence predictions far away from the
  training data and how to mitigate the problem.
\newblock In {\em Proceedings of the IEEE/CVF Conference on Computer Vision and
  Pattern Recognition}, pages 41--50, 2019.

\bibitem{hendrycks2019benchmarking}
Dan Hendrycks and Thomas Dietterich.
\newblock Benchmarking neural network robustness to common corruptions and
  perturbations.
\newblock {\em International Conference on Learning Representations (ICLR)},
  2019.

\bibitem{islam2021class}
Mobarakol Islam, Lalithkumar Seenivasan, Hongliang Ren, and Ben Glocker.
\newblock Class-distribution-aware calibration for long-tailed visual
  recognition.
\newblock {\em arXiv preprint arXiv:2109.05263}, 2021.

\bibitem{Ji2019BinwiseTS}
Byeongmoon Ji, Hyemin Jung, Jihyeun Yoon, Kyungyul Kim, and Younghak Shin.
\newblock Bin-wise temperature scaling (bts): Improvement in confidence
  calibration performance through simple scaling techniques.
\newblock {\em 2019 IEEE/CVF International Conference on Computer Vision
  Workshop (ICCVW)}, pages 4190--4196, 2019.

\bibitem{johnson2017driving}
Matthew Johnson-Roberson, Charles Barto, Rounak Mehta, Sharath~Nittur Sridhar,
  Karl Rosaen, and Ram Vasudevan.
\newblock Driving in the matrix: Can virtual worlds replace human-generated
  annotations for real world tasks?
\newblock In {\em 2017 IEEE International Conference on Robotics and Automation
  (ICRA)}, pages 746--753. IEEE, 2017.

\bibitem{karimi2022improving}
Davood Karimi and Ali Gholipour.
\newblock Improving calibration and out-of-distribution detection in deep
  models for medical image segmentation.
\newblock {\em IEEE Transactions on Artificial Intelligence}, 2022.

\bibitem{krishnan2020improving}
Ranganath Krishnan and Omesh Tickoo.
\newblock Improving model calibration with accuracy versus uncertainty
  optimization.
\newblock {\em Advances in Neural Information Processing Systems}, 2020.

\bibitem{kull2017beta}
Meelis Kull, Telmo Silva~Filho, and Peter Flach.
\newblock Beta calibration: a well-founded and easily implemented improvement
  on logistic calibration for binary classifiers.
\newblock In {\em Artificial Intelligence and Statistics}, pages 623--631.
  PMLR, 2017.

\bibitem{kumar2018trainable}
Aviral Kumar, Sunita Sarawagi, and Ujjwal Jain.
\newblock Trainable calibration measures for neural networks from kernel mean
  embeddings.
\newblock In {\em International Conference on Machine Learning}, pages
  2805--2814. PMLR, 2018.

\bibitem{kuppers2020multivariate}
Fabian Kuppers, Jan Kronenberger, Amirhossein Shantia, and Anselm Haselhoff.
\newblock Multivariate confidence calibration for object detection.
\newblock In {\em Proceedings of the IEEE/CVF Conference on Computer Vision and
  Pattern Recognition Workshops}, pages 326--327, 2020.

\bibitem{liang2020imporved}
Gongbo Liang, Yu Zhang, Xiaoqin Wang, and Nathan Jacobs.
\newblock Improved trainable calibration method for neural networks on medical
  imaging classification.
\newblock In {\em British Machine Vision Conference (BMVC)}, 2020.

\bibitem{lin2017focal}
Tsung-Yi Lin, Priya Goyal, Ross Girshick, Kaiming He, and Piotr Doll{\'a}r.
\newblock Focal loss for dense object detection.
\newblock In {\em Proceedings of the IEEE international conference on computer
  vision}, pages 2980--2988, 2017.

\bibitem{lin2014microsoft}
Tsung-Yi Lin, Michael Maire, Serge Belongie, James Hays, Pietro Perona, Deva
  Ramanan, Piotr Doll{\'a}r, and C~Lawrence Zitnick.
\newblock Microsoft coco: Common objects in context.
\newblock In {\em European conference on computer vision}, pages 740--755.
  Springer, 2014.

\bibitem{Liu_2022_CVPR}
Bingyuan Liu, Ismail Ben~Ayed, Adrian Galdran, and Jose Dolz.
\newblock The devil is in the margin: Margin-based label smoothing for network
  calibration.
\newblock In {\em Proceedings of the IEEE/CVF Conference on Computer Vision and
  Pattern Recognition (CVPR)}, pages 80--88, June 2022.

\bibitem{mukhoti2020calibrating}
Jishnu Mukhoti, Viveka Kulharia, Amartya Sanyal, Stuart Golodetz, Philip Torr,
  and Puneet Dokania.
\newblock Calibrating deep neural networks using focal loss.
\newblock {\em Advances in Neural Information Processing Systems},
  33:15288--15299, 2020.

\bibitem{ovadia2019can}
Yaniv Ovadia, Emily Fertig, Jie Ren, Zachary Nado, David Sculley, Sebastian
  Nowozin, Joshua Dillon, Balaji Lakshminarayanan, and Jasper Snoek.
\newblock Can you trust your model's uncertainty? evaluating predictive
  uncertainty under dataset shift.
\newblock {\em Advances in neural information processing systems}, 32, 2019.

\bibitem{platt1999probabilistic}
John Platt et~al.
\newblock Probabilistic outputs for support vector machines and comparisons to
  regularized likelihood methods.
\newblock {\em Advances in large margin classifiers}, 10(3):61--74, 1999.

\bibitem{ren2015faster}
Shaoqing Ren, Kaiming He, Ross Girshick, and Jian Sun.
\newblock Faster r-cnn: Towards real-time object detection with region proposal
  networks.
\newblock In {\em Advances in neural information processing systems}, pages
  91--99, 2015.

\bibitem{sakaridis2018semantic}
Christos Sakaridis, Dengxin Dai, and Luc Van~Gool.
\newblock Semantic foggy scene understanding with synthetic data.
\newblock {\em International Journal of Computer Vision}, 126(9):973--992,
  2018.

\bibitem{Objects3652019}
Shuai Shao, Zeming Li, Tianyuan Zhang, Chao Peng, Gang Yu, Xiangyu Zhang, Jing
  Li, and Jian Sun.
\newblock Objects365: A large-scale, high-quality dataset for object detection.
\newblock In {\em 2019 IEEE/CVF International Conference on Computer Vision
  (ICCV)}, pages 8429--8438, 2019.

\bibitem{sharma2017crowdsourcing}
Monika Sharma, Oindrila Saha, Anand Sriraman, Ramya Hebbalaguppe, Lovekesh Vig,
  and Shirish Karande.
\newblock Crowdsourcing for chromosome segmentation and deep classification.
\newblock In {\em Proceedings of the IEEE conference on computer vision and
  pattern recognition workshops}, pages 34--41, 2017.

\bibitem{simonyan2014very}
Karen Simonyan and Andrew Zisserman.
\newblock Very deep convolutional networks for large-scale image recognition.
\newblock {\em arXiv preprint arXiv:1409.1556}, 2014.

\bibitem{Segmenter2021}
Robin Strudel, Ricardo Garcia, Ivan Laptev, and Cordelia Schmid.
\newblock Segmenter: Transformer for semantic segmentation.
\newblock In {\em 2021 IEEE/CVF International Conference on Computer Vision
  (ICCV)}, pages 7242--7252, 2021.

\bibitem{szegedy2016rethinking}
Christian Szegedy, Vincent Vanhoucke, Sergey Ioffe, Jon Shlens, and Zbigniew
  Wojna.
\newblock Rethinking the inception architecture for computer vision.
\newblock In {\em Proceedings of the IEEE conference on computer vision and
  pattern recognition}, pages 2818--2826, 2016.

\bibitem{tian2019fcos}
Zhi Tian, Chunhua Shen, Hao Chen, and Tong He.
\newblock Fcos: Fully convolutional one-stage object detection.
\newblock In {\em Proceedings of the IEEE international conference on computer
  vision}, pages 9627--9636, 2019.

\bibitem{tomani2021post}
Christian Tomani, Sebastian Gruber, Muhammed~Ebrar Erdem, Daniel Cremers, and
  Florian Buettner.
\newblock Post-hoc uncertainty calibration for domain drift scenarios.
\newblock In {\em Proceedings of the IEEE/CVF Conference on Computer Vision and
  Pattern Recognition}, pages 10124--10132, 2021.

\bibitem{yu2020bdd100k}
Fisher Yu, Haofeng Chen, Xin Wang, Wenqi Xian, Yingying Chen, Fangchen Liu,
  Vashisht Madhavan, and Trevor Darrell.
\newblock Bdd100k: A diverse driving dataset for heterogeneous multitask
  learning.
\newblock In {\em Proceedings of the IEEE/CVF conference on computer vision and
  pattern recognition}, pages 2636--2645, 2020.

\bibitem{zhu2021deformable}
Xizhou Zhu, Weijie Su, Lewei Lu, Bin Li, Xiaogang Wang, and Jifeng Dai.
\newblock Deformable {\{}detr{\}}: Deformable transformers for end-to-end
  object detection.
\newblock In {\em International Conference on Learning Representations}, 2021.

\end{thebibliography}
}

\newpage

\twocolumn[\section*{\Large\textit{Supplementary Material:}  Bridging Precision and Confidence: A Train-Time Loss for Calibrating Object Detection}]

\section*{A. Overview}
\label{sec:over}
In this supplementary material, we provide the following items:

\begin{enumerate}[label={A.}{{\arabic*}},noitemsep,itemindent=0.5cm]
\item Further implementation details.
\item Additional experimental results.
\item More qualitative results.
\item Architecture Irrelevant: BPC with One-stage.
\item Formulation for Semantic Segmentation.
\item Error bars: Mean \& Std. Dev.
\item More results on Common Corruptions.
\end{enumerate}

\subsection*{A.1~Further Implementation Details}
\label{sec:impdet}
We choose Deformable-DETR (D-DETR) for our experiments. Our BPC loss is an auxiliary loss that jointly trains with the object detection losses to achieve better calibration. 
For our experiments, we use multi-gpu settings (4 GPUs) for training.

We provide further details on experimental settings for PascalVOC to watercolor1k, clipart1k, and comic1k domain shifts. We utilize train set of PascalVOC 2007 and 2012 for training and validation set of PascalVOC 2012 is used for evaluation purpose. For post-hoc method, it needs a hold-out validation set, and PascalVOC 2007 test set is used for that purpose. PascalVOC contains 20 categories of real images. Other out-domain datasets contain evaluation images, and respective 1k test set images are used for watercolor1k and comic1k, while whole 1k images of clipart1k are used for evaluation. Clipart1k also contains 20 categories, whereas 6 common categories are evaluated in watercolor1k and comic1k. We report calibration error (D-ECE) and mean average precision for reporting results.

\begin{table}[b]
\footnotesize
\centering
\renewcommand{\arraystretch}{1.1}
\begin{tabular}{lccc} 
\hline
\backslashbox{\textbf{Methods}}{\textbf{Scenario}}  & \multicolumn{3}{c}{\textbf{In-Domain (PascalVOC)}}                         \\ 
\hline\hline
                        & \textbf{D-ECE $\downarrow$} & \textbf{AP box} & \textbf{mAP@0.5}  \\ 
\textbf{Baseline \cite{zhu2021deformable}}       & 11.8   & 49.7                                                    & 73.8     \\ 
\textbf{TS (post-hoc) \cite{guo2017calibration}}             & 13.0     & 49.7                                                    & 73.8     \\ 
\textbf{MDCA \cite{hebbalaguppe2022stitch}}           & 12.8   & 48.9                                                    & 73.2     \\ 
\textbf{MbLS \cite{Liu_2022_CVPR}}           & 20.9   & 49.7                                                    & 73.6     \\ 
\hline
\textbf{\cellcolor[HTML]{D4F7D4}BPC (Ours)}           & \cellcolor[HTML]{D4F7D4}11.2   & \cellcolor[HTML]{D4F7D4}49.6                                                    & \cellcolor[HTML]{D4F7D4}74.0       \\
\hline
\end{tabular}
\captionsetup{justification=justified}
\caption{\small Calibration results with baseline, train-time losses and post-hoc methods are reported. BPC shows improvement in detection calibration for in-domain scenario. AP box and mAP@0.5 are also reported.}
\label{tab:pascalindomain}
\end{table}

\begin{table*}[t]
\footnotesize
\centering
\renewcommand{\arraystretch}{1.1}
\begin{tabular}{lccccccccc} 
\hline
\backslashbox{\textbf{Methods}}{\textbf{Scenarios}}         & \multicolumn{3}{c}{\textbf{Out-Domain (watercolor1k)}}                                                   & \multicolumn{3}{c}{\textbf{Out-Domain (clipart1k)}}                                                              & \multicolumn{3}{c}{\textbf{Out-Domain (comic1k)}}                                                         \\ 
\hline\hline
                  & \textbf{D-ECE $\downarrow$} & \textbf{AP box} & \textbf{mAP@0.5} & \textbf{D-ECE $\downarrow$} & \textbf{AP box} & \textbf{mAP@0.5} & \textbf{D-ECE $\downarrow$} & \textbf{AP box} & \textbf{mAP@0.5}  \\ 
\textbf{Baseline \cite{zhu2021deformable}} & 11.1           & 15.9                                                                      & 31.8             & 11.0             & 9.0                                                                         & 17.9             & 14.4           & 5.6                                                                       & 11.1              \\ 
\textbf{TS (post-hoc) \cite{guo2017calibration}}       & 19.3           & 15.9                                                                      & 31.8             & 19.2           & 9.0                                                                         & 17.9             & 22.0             & 5.6                                                                       & 11.1              \\ 
\textbf{MDCA \cite{hebbalaguppe2022stitch}}     & 9.8            & 17.5                                                                      & 34.8             & 10.5           & 9.3                                                                       & 17.8             & 15.3           & 4.9                                                                       & 8.8               \\ 
\textbf{MbLS \cite{Liu_2022_CVPR}}     & 16.3           & 16.8                                                                      & 34.5             & 12.3           & 9.2                                                                       & 17.9             & 17.7           & 5.5                                                                       & 10.4              \\ 
\hline
\textbf{\cellcolor[HTML]{D4F7D4}BPC (Ours)}     & \cellcolor[HTML]{D4F7D4}9.3            & \cellcolor[HTML]{D4F7D4}16.5                                                                      & \cellcolor[HTML]{D4F7D4}34.1             & \cellcolor[HTML]{D4F7D4}10.8           & \cellcolor[HTML]{D4F7D4}10.2                                                                      & \cellcolor[HTML]{D4F7D4}19.1             & \cellcolor[HTML]{D4F7D4}12.1           & \cellcolor[HTML]{D4F7D4}6.1                                                                       & \cellcolor[HTML]{D4F7D4}11.7              \\
\hline
\end{tabular}
\captionsetup{justification=justified}
\caption{\small Comparison of calibration performance with the baseline, train-time losses and post-hoc methods. Our BPC shows improvement over almost all competing approaches in three challenging out-domain scenarios i.e. from PASCALVOC to watercolor1k, clipart1k \& comic1k. AP box and mAP@0.5 are also reported for each scenario.}
\label{tab:pascaloutdomain}
\end{table*}

\begin{figure*}[t]
    \centering
    \includegraphics[width=0.85\linewidth]{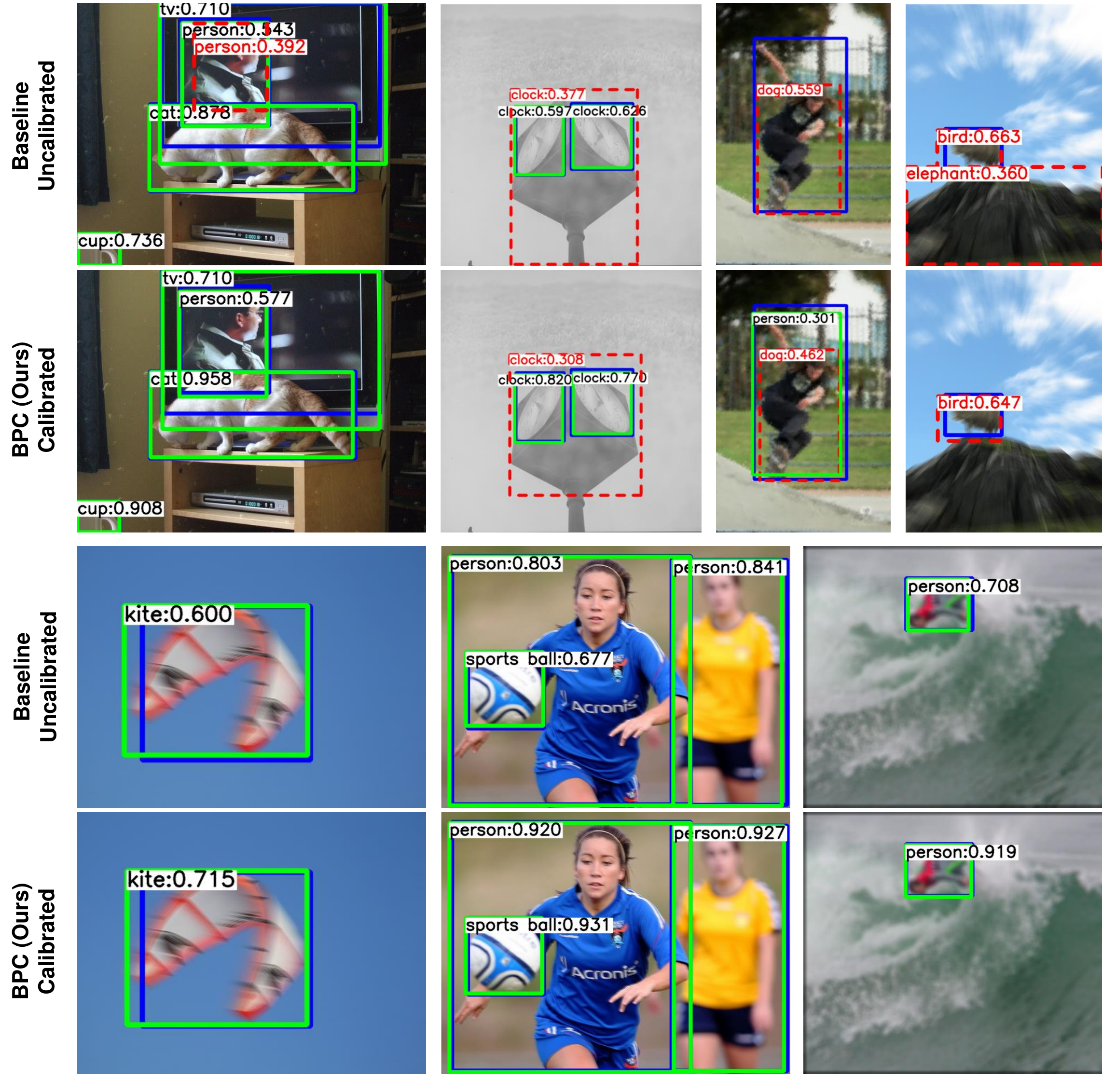}
    \captionsetup{justification=justified}
    \caption{\small Baseline \cite{zhu2021deformable} vs. BPC (Ours): Qualitative results on CorCOCO dataset (Out-Domain of MS-COCO). Detector trained with our loss forces the accurate predictions to be more confident whereas inaccurate predictions to be less confident. Detection threshold is set to \text{0.3}. \textcolor{green}{Green} boxes are accurate predictions with their respective confidence scores. \textcolor{red}{Red} (dashed) boxes are inaccurate predictions with corresponding scores. 
    \textcolor{blue}{Blue} shows the ground truth boxes for corresponding detections.}
    \label{fig:qual_COCO_outdomain}
    \vspace{-0.0cm}
\end{figure*}

\begin{figure*}[t]
    \centering
    \includegraphics[width=0.85\linewidth]{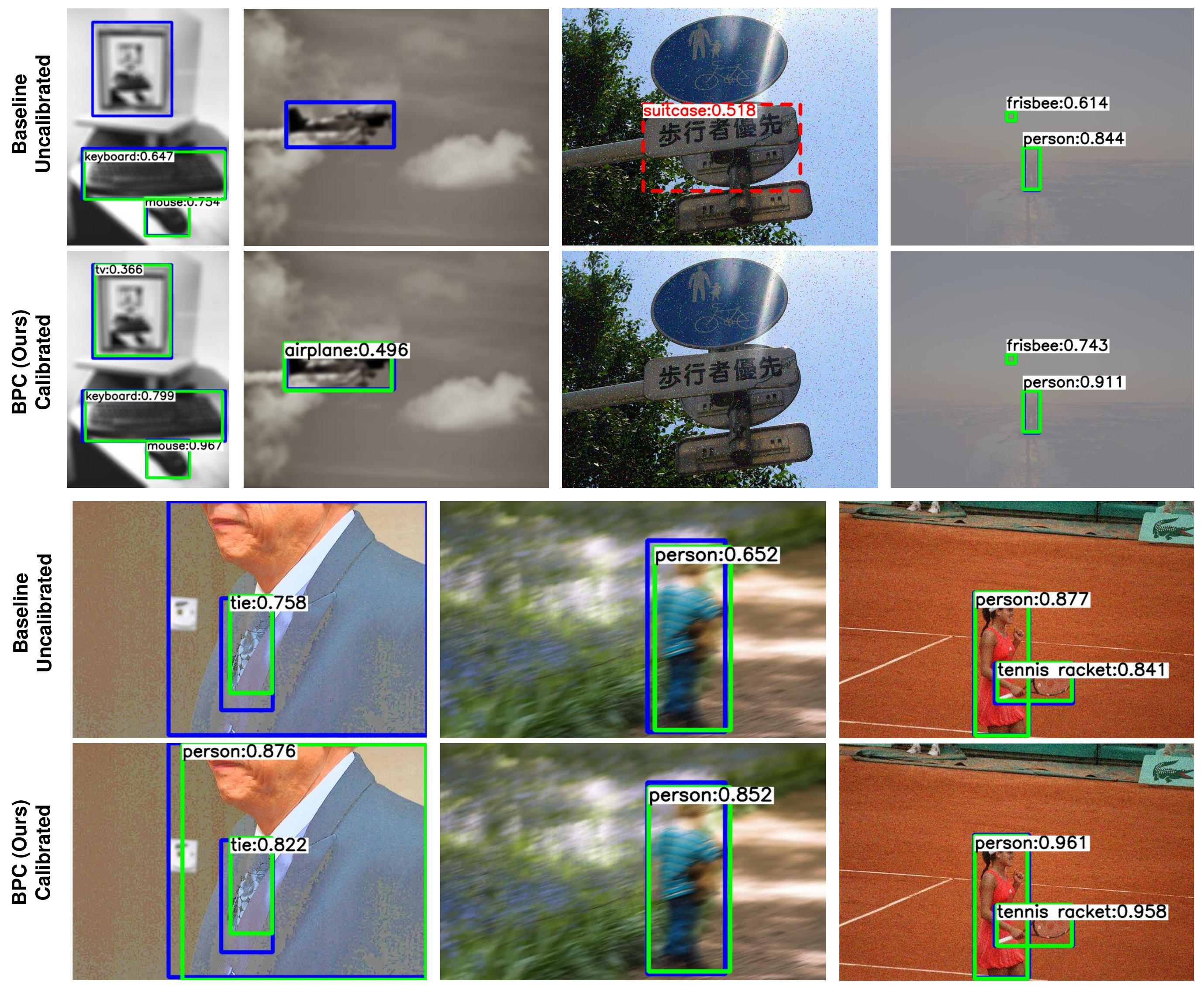}
    \captionsetup{justification=justified}
    \caption{\small Baseline \cite{zhu2021deformable} vs. BPC (Ours): Qualitative results on CorCOCO dataset (Out-Domain of MS-COCO). Detector trained with our loss forces the accurate predictions to be more confident whereas inaccurate predictions to be less confident. Detection threshold is set to \text{0.3}. \textcolor{green}{Green} boxes are accurate predictions with their respective confidence scores. \textcolor{red}{Red} (dashed) boxes are inaccurate predictions with corresponding scores. 
    \textcolor{blue}{Blue} shows the ground truth boxes for corresponding detections.}
    \label{fig:qual_COCO_outdomain2}
    \vspace{-0.0cm}
\end{figure*}

\subsection*{A.2~Additional Experimental Results}
\label{sec:pascal}
\textbf{PascalVOC:} We compare the performance of our calibration loss with recent calibration methods, post-hoc calibration method, and the baseline on PASCALVOC to watercolor1k, clipart1k, and comic1k domain shifts. In Tab. \ref{tab:pascalindomain}, we present the results on PascalVOC, and
our BPC loss reduces the D-ECE by 9.7\%$\downarrow$ over MbLS \cite{Liu_2022_CVPR} and 1.6\%$\downarrow$ over MDCA \cite{Liu_2022_CVPR}.

\noindent\textbf{Watercolor1k:} In Tab. \ref{tab:pascaloutdomain}, our BPC loss improves the calibration performance without losing significant detection accuracy. Our loss shows calibration improvement of 7.0\%$\downarrow$ over MbLS \cite{Liu_2022_CVPR} and 10.0\%$\downarrow$ over post-hoc method.

\noindent\textbf{Clipart1k:} Tab. \ref{tab:pascaloutdomain} shows that our BPC loss comparatively improves the calibration performance without losing detection accuracy. Our loss shows calibration improvement of 1.5\%$\downarrow$ over MbLS \cite{Liu_2022_CVPR} and 8.4\%$\downarrow$ over post-hoc method.

\noindent\textbf{Comic1k:} Our BPC loss improves the calibration performance along with the detection accuracy. In Tab. \ref{tab:pascaloutdomain}, our loss shows calibration improvement of 5.6\%$\downarrow$ over MbLS \cite{Liu_2022_CVPR} and 3.2\%$\downarrow$ over MDCA \cite{hebbalaguppe2022stitch}.

\subsection*{A.3~More Qualitative Results}
\label{sec:visual}
We show more qualitative calibration results on CorCOCO (corrupted version of MS-COCO 2017 validation set) in Figs. \ref{fig:qual_COCO_outdomain} and \ref{fig:qual_COCO_outdomain2}.
Detector trained with our loss forces the accurate predictions to be more confident whereas inaccurate predictions to be less confident.

\begin{table}[t]
\scriptsize
\centering
\renewcommand{\arraystretch}{1.1}
\tabcolsep=2.9pt\relax
\begin{tabular}{lcccccc}
\hline
Method & \multicolumn{3}{c}{\textbf{In-Domain (COCO)}}                                                        & \multicolumn{3}{c}{\textbf{Out-Domain (CorCOCO)}}                                                       \\ \hline
\textbf{}                  & \multicolumn{1}{c}{\textbf{D-ECE} $\downarrow$} & \multicolumn{1}{c}{\textbf{AP box}} & \textbf{mAP} & \multicolumn{1}{c}{\textbf{D-ECE} $\downarrow$} & \multicolumn{1}{c}{\textbf{AP box}} & \textbf{mAP} \\ 
\textbf{One-stage (FCOS)}          & \multicolumn{1}{c}{22.0}           & \multicolumn{1}{c}{38.7}            & 57.2             & \multicolumn{1}{c}{24.7}            & \multicolumn{1}{c}{20.4}            & 32.1      \\ \hline
\textbf{\cellcolor[HTML]{D4F7D4}BPC (Ours)}              & \multicolumn{1}{c}{\cellcolor[HTML]{D4F7D4}20.9}            & \multicolumn{1}{c}{\cellcolor[HTML]{D4F7D4}38.4}            & \cellcolor[HTML]{D4F7D4}56.9             & \multicolumn{1}{c}{\cellcolor[HTML]{D4F7D4}23.5}            & \multicolumn{1}{c}{\cellcolor[HTML]{D4F7D4}20.3}            & \cellcolor[HTML]{D4F7D4}32.0             \\ \hline
\end{tabular}
\captionsetup{justification=justified}
\caption{\footnotesize Calibration results on One-Stage object detector (FCOS). 
}
\label{tab:FCOSinoutreb}
\end{table}

\subsection*{A.4~Architecture Irrelevant: BPC with One-stage}
\label{sec:archirr}
In Tab.~\ref{tab:FCOSinoutreb}, we show results with a one-stage detector (FCOS \cite{tian2019fcos} with ResNet-50). Our BPC loss improves calibration in both in-domain (COCO) and out-domain (CorCOCO).

\begin{figure}[t]
    \centering
    \includegraphics[width=0.8\linewidth]{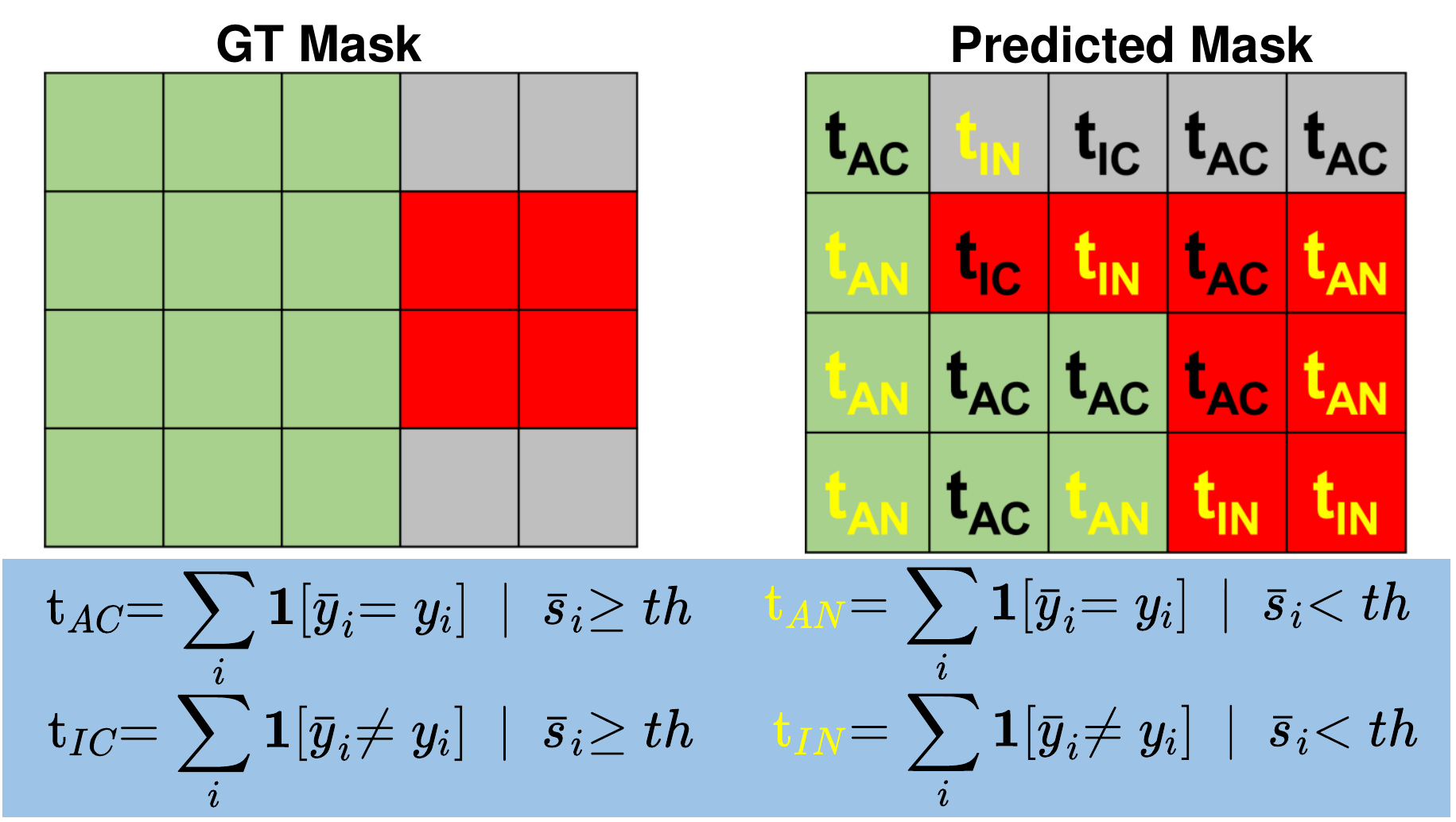}
    \captionsetup{justification=justified}
    \caption{\footnotesize BPC loss for Semantic Segmentation task. 3 classes as box colors (green, red \& grey). Black text in box (confident) and yellow (not-confident) predictions.} 
    \label{fig:semantic}
\end{figure}

\subsection*{A.5~Formulation for Semantic Segmentation}
\label{sec:semseg}
Our loss BPC is extensible for semantic segmentation. We provide a sketch formulation to extend BPC loss for semantic segmentation (Fig.~\ref{fig:semantic}). 

\subsection*{A.6~Error bars: Mean \& Std. Deviation}
\label{sec:errbars}
\begin{figure*}[t]
    \centering
    \includegraphics[width=0.75\linewidth]{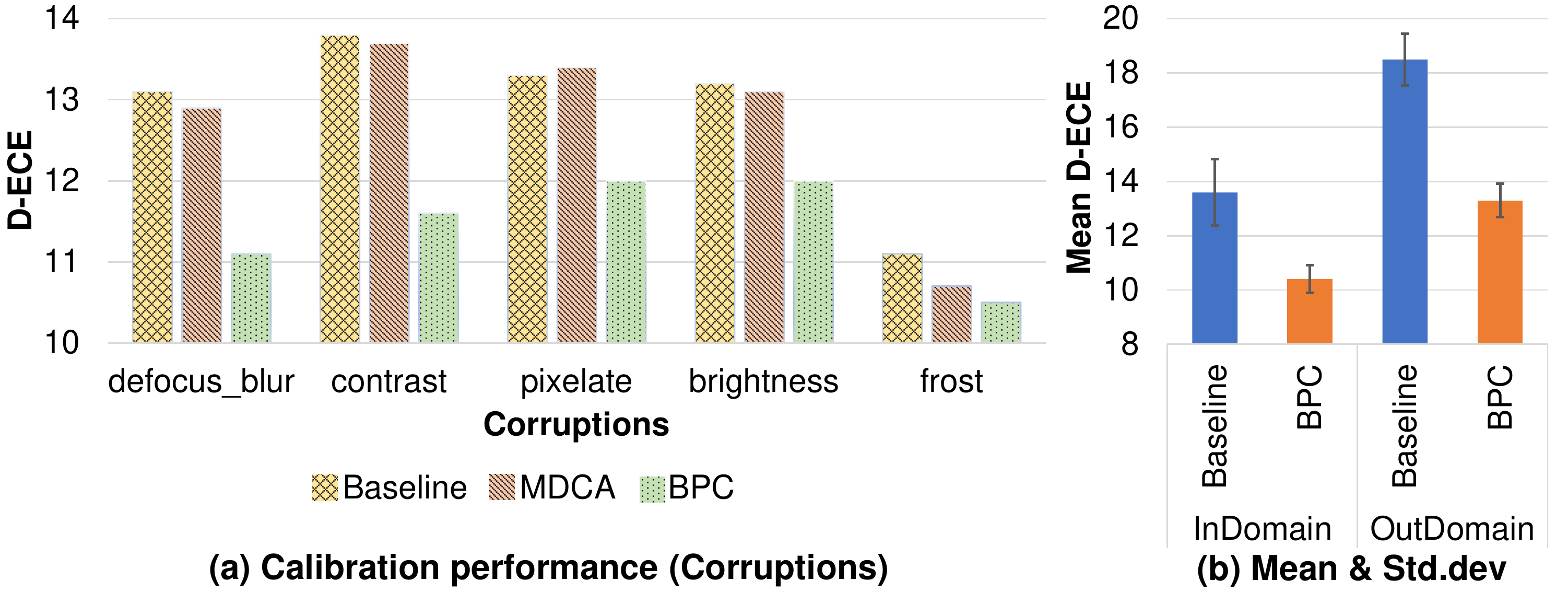}
    \captionsetup{justification=justified}\vspace{-0.8em}
    \caption{\footnotesize(a) Calibration performance on COCO corruptions. (b) Error bars on CS/Foggy-CS dataset.}. 
    \label{fig:corerror}
    \vspace{-2.0em}
\end{figure*}

Fig.~\ref{fig:corerror}(b) plots the mean and std.dev D-ECE for BPC and baseline in CS (in-domain) and Foggy-CS (out-domain).

\subsection*{A.7~More results on Common Corruptions}
\label{sec:comcor}
We corrupt COCO evaluation set (val2017) after sampling 5 different corruptions with a fixed severity level of 2 from Common Corruptions \cite{hendrycks2019benchmarking}. 
Fig.~\ref{fig:corerror}(a) shows that our BPC loss can improve the calibration performance on all five corruptions.

\end{document}